\documentclass[pre,twocolumn,twoside,byrevtex,superscriptaddress,floatfix]{revtex4-1}

\usepackage[realmainfile]{currfile}

\usepackage[caption=false]{subfig}
\usepackage[table]{xcolor}
\usepackage{sidecap}
\usepackage{float}
\usepackage{colortbl}
\usepackage{array}

\usepackage{color}

\makeatletter

\def\CT@@do@color{%
  \global\let\CT@do@color\relax
  \@tempdima\wd\z@
  \advance\@tempdima\@tempdimb
  \advance\@tempdima\@tempdimc
  \advance\@tempdimb\tabcolsep
  \advance\@tempdimc\tabcolsep
  \advance\@tempdima2\tabcolsep
  \kern-\@tempdimb
  \leaders\vrule
  \hskip\@tempdima\@plus  1fill
  \kern-\@tempdimc
  \hskip-\wd\z@ \@plus -1fill }
\makeatother

\usepackage[table]{xcolor}

\usepackage{graphicx,epsfig,verbatim,enumerate}
\usepackage{amssymb,amsmath}
\usepackage{ifthen}

\usepackage{longtable}

\usepackage{mathtools}

\newboolean{twocolswitch}

\newcommand{\sindex}[1]{}
\newcommand{\nindex}[1]{}

\newcommand{\www}[1]{\url{#1}}

\usepackage{lettrine}

\setboolean{twocolswitch}{true}

\begin{document}

\title{\protect
Computational Paremiology:
Charting the temporal, ecological dynamics of proverb use in books, news articles, and tweets
}

\author{
\firstname{Ethan}
\surname{Davis}
}
\email{ethan.davis@uvm.edu}

\affiliation{
  Computational Story Lab,
  Vermont Complex Systems Center,
  MassMutual Center of Excellence for Complex Systems and Data Science,
  Vermont Advanced Computing Core,
  University of Vermont,
  Burlington, VT 05401.
  }

\author{
\firstname{Christopher M.}
\surname{Danforth}
}
\email{chris.danforth@uvm.edu}

\affiliation{
  Computational Story Lab,
  Vermont Complex Systems Center,
  MassMutual Center of Excellence for Complex Systems and Data Science,
  Vermont Advanced Computing Core,
  University of Vermont,
  Burlington, VT 05401.
  }

\affiliation{
  Department of Mathematics \& Statistics,
  University of Vermont,
  Burlington, VT 05401.
  }

\author{
\firstname{Wolfgang}
\surname{Mieder}
}
\email{wolfgang.mieder@uvm.edu}

\affiliation{
  Department of German \& Russian,
  University of Vermont,
  Burlington, VT 05401.
  }

\author{
\firstname{Peter Sheridan}
\surname{Dodds}
}
\email{peter.dodds@uvm.edu}

\affiliation{
   Computational Story Lab,
  Vermont Complex Systems Center,
  MassMutual Center of Excellence for Complex Systems and Data Science,
  Vermont Advanced Computing Core,
  University of Vermont,
  Burlington, VT 05401.
}

\affiliation{
  Department of Computer Science,
  University of Vermont,
  Burlington, VT 05401.
  }

\date{\today}

\begin{abstract}
  \protect
  Proverbs are an essential component of language and culture, and though much attention has been paid to their history and currency, there has been comparatively little quantitative work on changes in the frequency with which they are used over time. 
With wider availability of large corpora reflecting many diverse genres of documents, it is now possible to take a broad and dynamic view of the importance of the proverb. 
Here, we measure temporal changes in the relevance of proverbs within three 
corpora, differing in kind, scale, and time frame:
Millions of books over centuries;
hundreds of millions of news articles over twenty years;
and 
billions of tweets over a decade. 
We find that proverbs 
present heavy-tailed frequency-of-usage rank distributions in each venue;
exhibit trends reflecting the cultural dynamics of the eras covered; 
and 
have evolved into contemporary forms on social media.

\end{abstract}


\maketitle

\section{Introduction}
\label{sec:introduction}

Our goal here is to advance `computational paremiology':
The data-driven study of proverbs.
We first build a quantitative foundation 
by estimating the frequency of use over time
for an ecology of proverbs in several large corpora from different domains.
We then characterize basic temporal dynamics allowing us
to address fundamental questions such as 
whether or not 
proverbs appear in texts 
according to a similar distribution to words \cite{zipf1949a, balasubrahmanyan_quantitative_1996, ryland_williams_zipfs_2015, ferrer_i_cancho_two_2001}.

In studies of phraseology, data on frequency of use is often conspicuously absent \cite{cermak_proverbs_2014}. 
The recent proliferation of large machine-readable corpora has enabled new frequency-informed studies of words and $n$-grams that have expanded our knowledge of language use in a variety of settings, from the Google Books $n$-gram Corpus and the introduction of ``culturomics'' \cite{michel_quantitative_2011,pechenick_characterizing_2015}, 
to availability and analysis of Twitter data \cite{alshaabi_storywrangler_2020}.  
However, routine formulae, or multi-word expressions that cannot be reduced to a literal reading of their semantic components, remain notoriously averse to reliable identification despite carrying high degrees of symbolic and indexical meaning \cite{goos_multiword_2002}. 
It is, for instance, much easier to chart a probability distribution of single words or $n$-grams than complex lexicon-dependent utterances such as proverbs, conventional metaphors, or idioms.


Perhaps the most recognizable routine formulae are proverbs and their close cousin, idioms. Centuries of the study of proverbs---paremiology---have shown their importance in language and culture, and that they are immensely popular among the folk \cite{mieder_proverbs_2012}. 
Proverbs are generally metaphorical in their use, and map a generic situation described by the proverb to an immediate context. 
In light of challenges in developing reliable instruments for measurement and quantification of figurative language, research would greatly benefit, as it has with words, from a better understanding of the frequency and dynamics of proverb use in texts. By applying new methodologies in measuring frequency and probability distributions, this study seeks to contribute to this endeavor.

Before going any further,
we must detail a more precise definition of the proverb. 
Though there is still some debate, it is widely agreed that proverbs are popular sayings that offer general advice or wisdom. 
However, naturally not all such sayings are proverbs. 
Many attempts at more precise definitions have been made, perhaps simplest being that of Gallacher: ``A proverb is a concise statement of an apparent truth which has [had, or will have] currency among the people.'' 
This definition, while convenient, leaves out some important features, like their metaphoricity, and their dependence on context \cite{mieder_proverbs_2008}.

Mieder's definition is perhaps the most useful for our present purposes: ``Proverbs [are] concise traditional statements of apparent truths with currency among the folk. More elaborately stated, proverbs are short, generally known sentences of the folk that contain wisdom, truths, morals, and traditional views in a metaphorical, fixed, and memorizable form and that are handed down from generation to generation'' \cite{mieder_proverbs_2008}.

Proverbs maintain a particular relationship with their context of use that provides a fruitful domain for frequency and probability analysis. 
An important part of the proverb is the context in which it is used. The metaphorical property of a proverb need not only have to do with the proverb itself (as in the proverb/metaphor ``war is hell'', in which war is compared to hell within the proverb). 
In general, the use of a proverb is metaphorical in context, meaning that the proverb offers wisdom about a current situation via a metaphoric comparison to a proverbial one \cite{mieder_proverbs_2008}. 
For instance, while the proverb ``still waters run deep'' 
might be 
used to caution someone against taking a seeming calm for granted, as it may belie unseen dangers. 
As with many other proverbs, it is hard to imagine anyone using the proverb ``you can't put lipstick on a pig'' in any literal or pragmatic context. Rather, these phrases offer wisdom embodied in the culture as opposed to that of the speaker. In this way proverbs may be used generically without proffering personal expertise. 

Indeed, proverbs are necessarily ambiguous enough to offer wisdom in any number of situations. 
Michael Lieber argued that this ambiguity paradoxically gives proverbs the function of disambiguating situations in which they are used. In part due to their role as cultural rather than individual wisdom, they can be invoked impersonally as a way of clarifying a complex reality \cite{mieder_analogic_1994}. As such, part of Winick's definition of the proverb is that they ``address recurrent social situations in a strategic way'' \cite{mieder_proverbs_2008}.

It is important to note the distinction between proverbs and idioms. An example of an idiom would be the phrase ``red herring'' denoting a mislead.
The meanings of idioms, like proverbs, often cannot be ascertained from the meanings of their component words. But unlike proverbs, idioms are often not complete sentences, require context, and need not reference a paradigmatic situation. 
Proverbs on the other hand represent a complete situation and offer some sort of general wisdom. 
The boundary between the two however is rather fuzzy and contains many idioms and proverbial expressions. 
For instance the proverb ``every cloud has its silver lining'' is perhaps more well known by its idiomatic reduction ``silver lining''. 
In fact, people may use an idiom 
without any knowledge of the its proverbial context. 
Our intent here is to focus
on expressions of full proverbs, and not their idiomatic uses. 
As previous work has shown, 
it is possible to investigate the manipulations and idiomizations of individual proverbs \cite{cermak_proverbs_2014, moon_fixed_1998}, 
and part of our 
study is devoted to continuing that work. 
However, our approach certainly has limitations,
and further research into flexible searches 
or other identification methods will be essential in future work.  

Metaphor and idiom identification and comprehension are an open area of research in machine learning and NLP (Natural Language Processing) \cite{fazly_unsupervised_2009, shutova-2010-models}. In general, metaphors and metaphorical speech are difficult to identify, and do not occur in consistent repeated phrasings. Whereas in ``bag-of-words'' methods, one is allowed the tacit assumption that most of these words are represented in the lexicon of the language in the search for routine formulae, one must access the lexicon as an essential step in verifying a phrase's meaningfulness. Furthermore, the source and target domains of their mapping are seldom explicit,
as laid out by Lakoff and Johnson in their Conceptual Metaphor Theory
\cite{lakoff_metaphors_1985, andersson_understanding_2013}. 
However, proverbs generally appear in the same recognizable format, and in the form of a full, self-contained sentence. 
Prospectively, understanding of the conceptual mapping involved in proverb use may provide a useful step towards general understanding of metaphors in the above fields \cite{ozbal_learning_2016, ozbal_prometheus_2016}.

Arguably, the proverb's flexibility of use has helped make them
an essential part of language and communication, literature, discourse, and media \cite{mieder_proverbs_2012}. Interest in the collection and study of proverbs dates back to at least the ancient Greeks and Sumerians. Erasmus famously collected proverbs. In English literature, the proverb has been an important device for many famous authors, among them Geoffrey Chaucer, William Shakespeare, Oscar Wilde, and Agatha Christie \cite{mieder_literary_1994, mieder_proverbs_1994}.

Modern politics attests to the continued relevance of the proverb. 
In politics, proverbs have been employed as a way to communicate succinctly and persuasively with the populace. Early American politicians like Benjamin Franklin used proverbs to help shape a national identity and character, as with his still widely read/cited \textit{Poor Richard's Almanac}. Abraham Lincoln employed proverbs in his famous speeches surrounding the American Civil War and Emancipation. During the Second World War, Churchill, Truman, and Hitler all famously used proverbs in their speeches and slogans \cite{mieder_proverbs_2005}. During Emancipation and the American civil rights movement respectively, proverbs were used by Frederick Douglass, and Martin Luther King Jr., to motivate the people and communicate moral values \cite{mieder_proverbs_2005}. 
More recently, dominant political figures in the US 
like Barack Obama and Bernie Sanders have used
proverbs to great effect \cite{mieder_right_2019}, 
and political and religious interests 
try to shape which proverbs children are taught in school \cite{lau_cheaters_2004}.

\subsection{Quantitative approaches}

This is by no means the first quantitative study of proverb use. Permiakov called for demographic studies of proverb knowledge to gather an impression of which proverbs were being used by the folk, in the interest of establishing a paremiological minimum: A minimum lexicon of proverbs for a language \cite{permiakov_question_1989}. Subsequent interest in proverb knowledge in psychology and folklore resulted in several studies conducted in the United States. 
Early studies by Albig and Bain in the 1930s found that American college students could recall on average between 25 and 27 distinct proverbs, many of which were common among participants \cite{albig_proverbs_1931, bain_verbal_1939}. 
A more recent study by Haas observed proverb familiarity among college students in several regions of the US. 
They performed experiments in both proverb generation and proverb recognition. Notably, students could recognize more proverbs than they could recall on their own \cite{haas_proverb_2008}.

Apart from the lexicographic collection of proverbs from texts, several attempts have been made to quantify and characterize their use. 
Whiting, in his assiduous collection of proverbs from texts in ``Modern Proverbs and Proverbial Sayings'' \cite{whiting_modern_2014}, kept track of the frequency with which they were encountered. 
Norrick attempted a manual search for proverb frequency, 
though he was constrained to only using proverbs starting with the letter f, and used a relatively small text sample \cite{norrick_how_1985}. 
In the first serious computational analysis of proverb frequency, 
Lau searched for and counted 
instances of proverbs in newspapers in the Lexis/Nexis ALLNWS database \cite{lau_its_1996}.

David Cram theorized that proverbs, acting as self-contained lexical units, were employed much in the same way that words are, and that their use involved a ``lexical loop'' where the speaker accesses the lexicon in addition to the syntax when forming a text. 
As such, in the case of proverbs (and phrasal idioms), 
one ought to ``analyze a syntactic string as a single lexical item'' \cite{mieder_linguistic_1994}.

Moon's exhaustive early study of fixed expressions and idioms (denoted FEIs)  in the Oxford Hector Pilot Corpus (OHPC) did just that \cite{moon_fixed_1998}. His study represents the first serious attempt to apply the new tools of computational linguistics to routine formulae. He searched the OHPC (a precursor to the British National Corpus or BNC) for instances of 6776 FEIs from the Collins Cobuild English Language Dictionary. It is worth noting that at the time, there were few machine-readable English phraseological lexica. Though proverbs consisted of only 3.5\% of the searched phrases (240), 19\% of the expressions found in the corpus were proverbial expressions, the second most common subtype behind ``simple expressions'' (70\%). Of the proverbs found, 59\% were deemed metaphorical. Moon notes that exploitation of FEIs are easy to miss, and uses the proverb ``a bird in the hand is worth two in the bush'' as an example.

Significantly, Moon noted that journalism was over-represented in the corpus, and that the results did not represent the distributions of these FEIs in English as a whole. This and other similar caveats inspired the present study to observe genre-specific corpora separately, and compare after analysis.

\begin{figure*}[tp!]
    \centering
    \includegraphics[width=\textwidth]{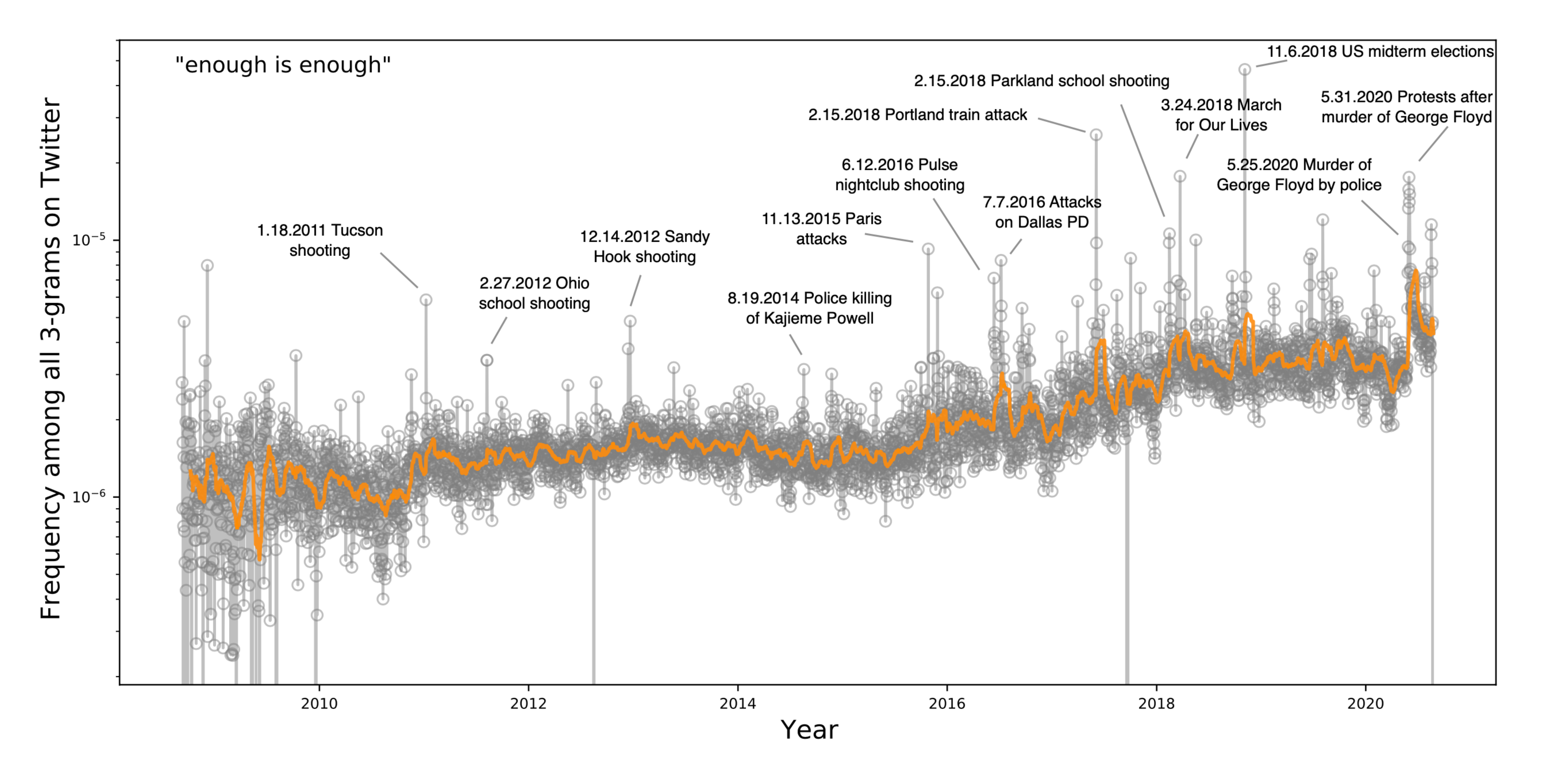}
    \caption{
        \textbf{Daily relative frequency of the 3-gram ``enough is enough'' on Twitter}. 
        The popularity of ``enough is enough'' on Twitter grew steadily over the last decade, and it has been the most popular proverb on Twitter since 2016, perhaps originating from its consistent use by Senator Bernie Sanders \cite{mieder_right_2019}. It has since become associated with growing protests against police brutality and gun violence. Annotations reflect widely reported violent events and protests (with the exception of the 2018 US midterm elections). The stark simplicity of this sixteenth century proverb evokes a narrative of repetition past the point of tolerance~\cite{mieder_dictionary_1992}. In this instance, beginning as a condemnation of the continued reaffirmation of the status quo in US politics by Senator Sanders, it is now popular as collective outcry against political inaction in the wake of regular mass shootings in the US, and a lack of accountability in the killing of black Americans by police. The changing significance and popularity of the proverb in the past decade displays the aptitude of proverbial speech to be successfully employed in varying contexts, and its potential to illustrate narrative commonalities between phenomena.
        }
\end{figure*}

\v{C}erm\'{a}k's essay collection ``Proverbs: Their Lexical and Semantic Features'' contains several essays that deal with the distribution of proverbs in the British National Corpus \cite{cermak_proverbs_2014}. In \v{C}erm\'{a}k's pioneering essays, he searches for occurrences of English proverbs in the BNC corpus (100 million words) \cite{BNC}. In this study, even the most common proverbs seem to occur relatively infrequently. For example, ``easier said than done'' is the most common, appearing 62 times in the entire corpus. His study discusses the relevance of corpus occurrence to a paremiological minimum (He uses a limited proverb list from Wiktionary). Another study focuses on text introducers to various proverbs using collocation analysis. (\v{C}erm\'{a}k notably created/spearheaded one of the first machine-readable phrasaeological lexica in the ``Czech Idiom Dictionary'' (1994).)

\v{C}erm\'{a}k relates frequency dictionaries to discussions of a paremeological minimum. Should proverb frequency in large corpora be taken into account when judging that minimum? Of course, there are problems with this approach as well: proverbs rely heavily on oral tradition, and are prone to frequent corruptions and purposeful exploitations. As such there is no guarantee that a search of a given phrasing of a proverb will capture all, if any, of its occurrences in a text. There are ways around this on an individual basis, but it depends on the proverb: some employ parallel structures (like ``good X make good Y''), or have popular idiomizations (like ``silver lining''). 

Most recently, in an introductory paremiology textbook \cite{hrisztova-gotthardt_proverbs_2015},
Steyer outlined a process general corpus linguistic method for studying proverbs, similar to Moon and \v{C}erm\'{a}k. 
Here, we expand on the above literature, including much larger corpora and proverb data sets. 

Should the ambition be to find these distributions in English as a whole?
We contend that there is no such universal corpus for any language.
Clearly use of these phrases is context-dependent, it seems unlikely inter-contextual searches will yield greater insight than single-genre searches. 
Instead, frequency dynamics and distributions in separate corpora from differing contexts may be more informative. 

\subsection{From Data on Language to Culture}

For our present study of proverbs from a corpus linguistic point of view,
we focus on two problems, which we may frame, however artificially, as questions:
1) How does frequency of proverb use compare across proverbs, and does that distribution echo previous findings in linguistics? 
and 
2) What stories emerge once the dimension of time is added to our observations of the frequency of proverb use in these corpora? 
Can shifts in popularity be related to known events, 
and can our knowledge of the history of proverb 
use be advanced through these methods?

One of the foundational achievements in the study of complex systems was Zipf's identification of scaling laws in language and other social phenomena~\cite{zipf_human_2012}. 
Indeed as early as 1996, natural language (in the context of computational linguistics) was cited explicitly as an example of the recently coined ``complex adaptive systems'' \cite{balasubrahmanyan_quantitative_1996}. 
It was first observed by Zipf that the rank distribution of words in a text follows a power law $F(r) = cr^{-\alpha}$, where $r$ is a word's rank, $F(r)$ is its frequency,  with $\alpha \simeq 1$. While primary interest here is paid to its appearance and seeming ubiquity in language, the same class of distributions have been observed in phenomena across a wide range of fields including physics, biology, psychology, sociology, urban studies, and engineering \cite{clauset_power-law_2009, martinez-mekler_universality_2009}.

Several studies have addressed possible mechanisms for the emergence of these distributions from empirical data. Notably, work by Dodds et al.\ showed that the distribution results from a Simon competition model, in which the first mover has an advantage \cite{dodds_simons_2017}. In this case the older proverbs may have a competitive edge in their proliferation and popularity. Cancho et al.\ showed in a language generating genetic algorithm that optimal results for both low speaker and receiver effort followed a Zipf distribution \cite{cancho_least_2003}. 

 While Zipf observed this phenomenon for words in a text, it has since been observed that individual words in a large corpus follow a broken power law distribution and do not strictly adhere to Zipf's law \cite{ryland_williams_zipfs_2015}.  Several attempts have been made to generalize the original Zipf distribution. Benoit Mandelbrot derived an analogous distribution using information theory, dubbed the Zipf-Mandelbrot distribution \cite{mandelbrot_informational_1953}. More recently, Cancho and Sole formalized a broken power law distribution with two distinct scaling regimes \cite{ferrer_i_cancho_two_2001}.


\begin{figure*}
    \centering
    \includegraphics[width=\textwidth]{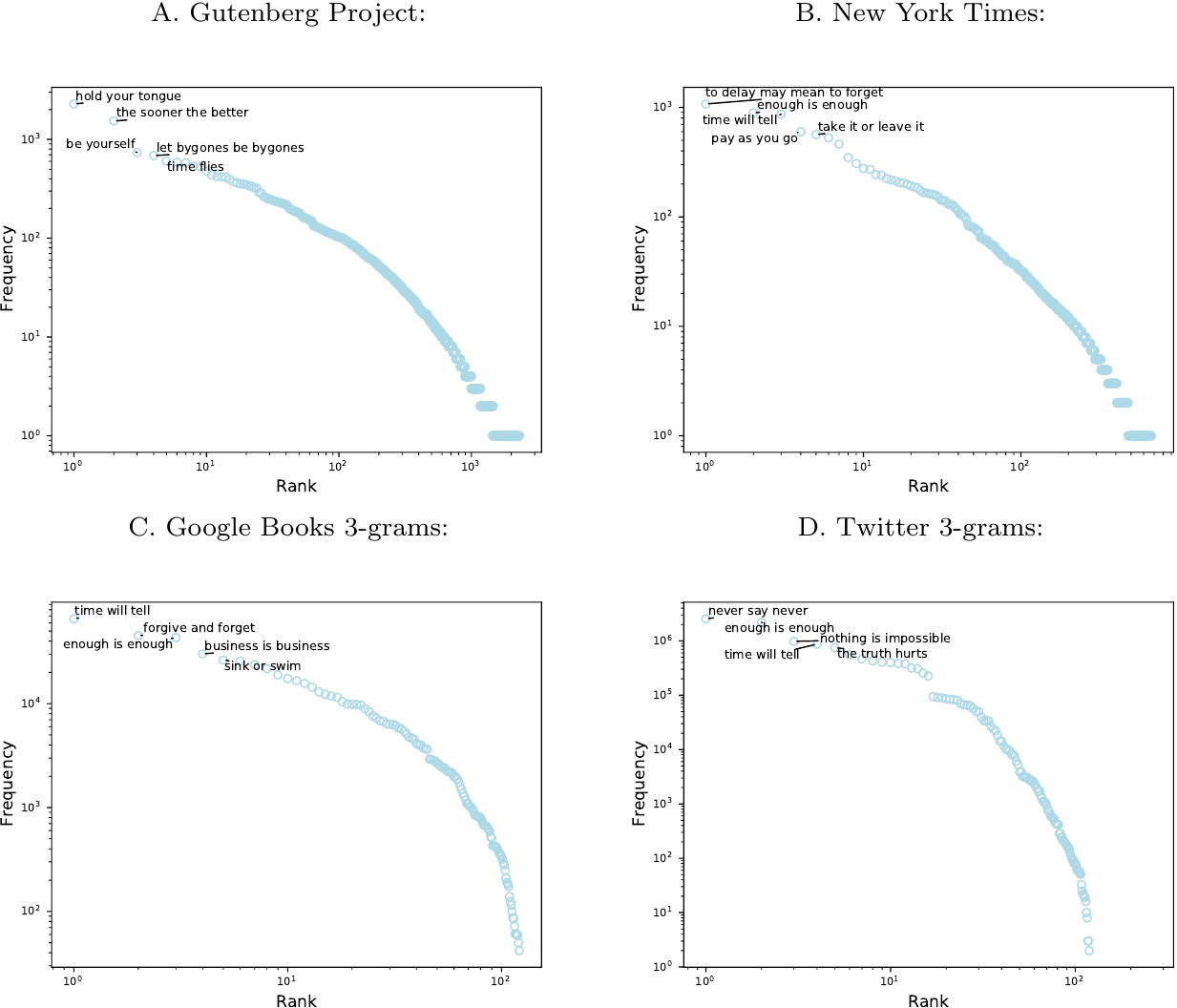}
    \caption{\textbf{Zipf distributions for entries from Mieder's \textit{Dictionary of American Proverbs} \cite{mieder_dictionary_1992}.} For each corpus, proverbs are enumerated and shown on logarithmic axes as a function of rank, with `hold your tongue', `to delay may mean to forget', `time will tell', and `never say never' topping the charts in Gutenberg, NYT, Google, and Twitter respectively. Each distribution exhibits heavy-tailed behavior, more prominently for Gutenberg and NYT .}
    \label{fig:paremiology.zipf}
\end{figure*}

One shortcoming noted in many evaluations of Zipf's law in text is that power law scaling breaks down toward the tail of these empirical distributions. Recent work by Williams et al.\ \cite{ryland_williams_zipfs_2015} however, showed that power law scaling holds over more orders of magnitude when randomly partitioned phrases are used rather than individual words. That study also suggested a refocusing of corpus linguistic attention from words to phrases as essential elements of language. Further work by Williams et al.\ \cite{williams_text_2015} suggested that changes in scaling in Zipf distrubutions of large corpora can be attributed to text mining. Few, if any, attempts have been made to apply Zipf's law to phraseological lexica. 

With large amounts of newly digitized text, corpus linguistics and lexicology/lexicography have seen renewed wider interest, and new results. Can these methods be used to tell new stories that are of interest to those working in the humanities? And in particular, how can that work embed itself into the existing wealth of knowledge accrued by those disciplines. In this case, how can computational work on proverbs situate itself in the existing knowledge-base of paremiology?

\begin{figure*}[tp!]
    \centering
    \includegraphics[width=\textwidth]{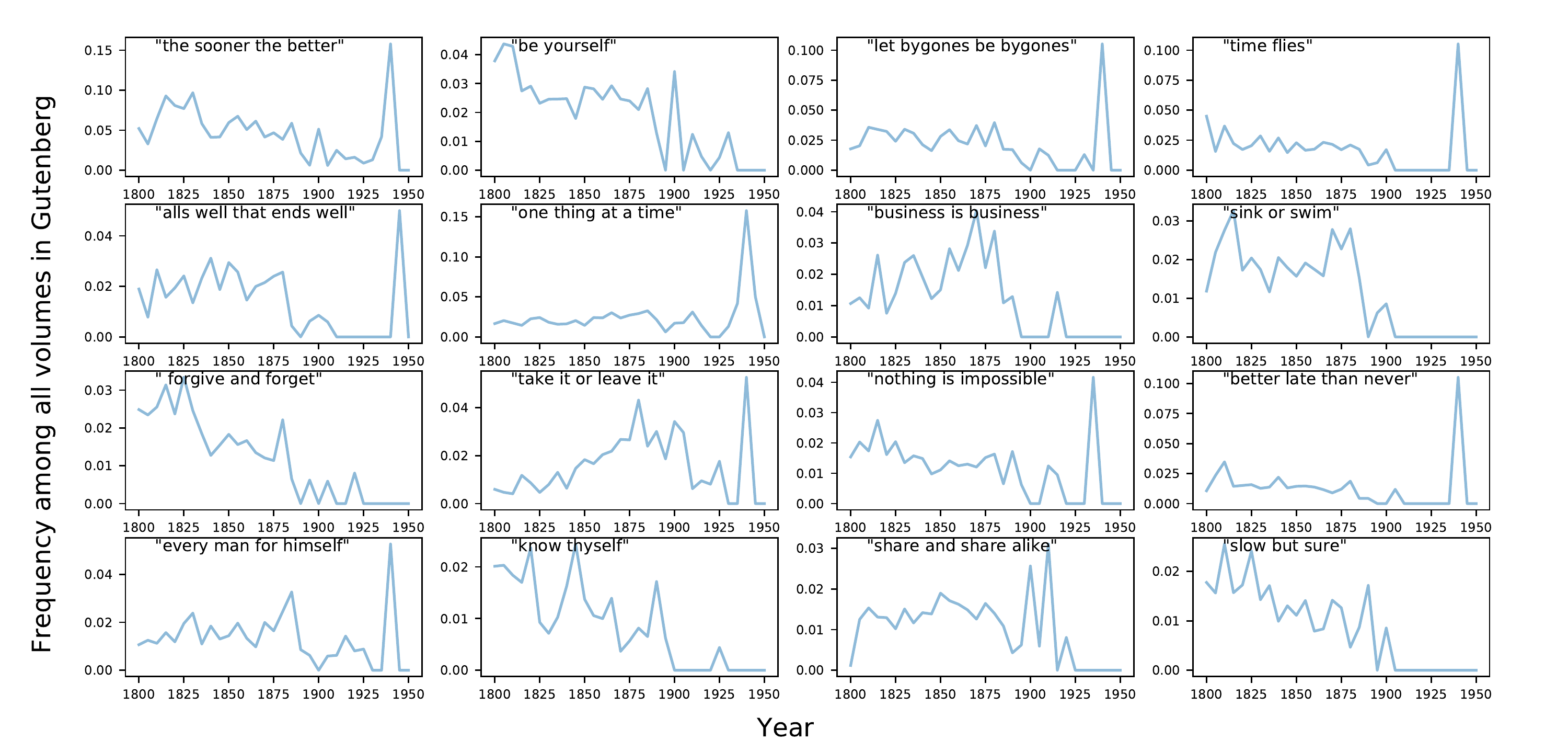}
    \caption{\textbf{Time series for the 16 most popular proverbs in the Gutenberg corpus, ranked by overall count.}
    These most common proverbs occur in a large portion of documents in the corpus for most of the period studied. 
    For instance, ``the sooner the better'' regularly appeared in at least 5\% of documents from the 19th century. 
    Plots are ordered in the grid by rank first left to right, then top to bottom. 
    Note that the vertical axis ranges vary across plots to highlight individual variation in time.
    }
\end{figure*}

In their seminal 2011 paper, Michel et al.\ discussed the newly created Google Books corpus, and coined the term ``culturomics'' to describe the nascent discipline concerned with observable trends in the use of $n$-grams over time \cite{michel_quantitative_2011}. They present several convincing case studies, among them trends in the use of ``influenza'' with historical outbreaks, and the use of geographical and antagonistic terms alongside the history of the American Civil War. These case studies make use of time series data and relative frequency to tell complex stories of interest from simple queries. 

However, Pechenick et al.\ note that there are issues with Google Books' representation of culture.
For one, books are not indexed by popularity, and each book appears only once. As a result, the linguistic contributions of the most popular books are weighted equally with the least popular \cite{pechenick_characterizing_2015}. 
Secondly,
the increase in volume of scientific publications in the last century causes the last century of English as a whole to be relatively skewed towards that genre. 
For instance, enormously  influential books like \textit{To Kill a Mockingbird}, \textit{I Know Why the Caged Bird Sings}, \textit{Mockingjay}, or \textit{Harry Potter and the Order of the Phoenix} are only represented once, and share the same weight as any other book 
(new editions notwithstanding). 
In the last century, the rise in volume of scientific and academic publication drastically increased the relative influence of this type of writing. 
here, we examine only the English Fiction subset of the corpus,
which can be partially defended~\cite{pechenick2015b}. 

Other work by Reagan et al.\ utilized the timelines \textit{within} texts to evaluate the emotional arc of a text, given word valence (sentiment) data. Inspired by Kurt Vonnegut's rejected Master's thesis (in Anthropology) on the shapes of stories, they found that indeed the emotional arcs of most stories in the Gutenberg corpus could be reduced to a handful of paradigmatic shapes \cite{reagan_emotional_2016}.

Work by Underwood et al.\ used historical use of gendered names and words to reveal trends in gender representation in literature using data from the HathiTrust digital library \cite{underwood_transformation_nodate}. 

StoryWrangler, a tool recently developed by Alshaabi et al.\ allows users to explore the temporal dynamics of $n$-grams found on Twitter \cite{alshaabi_storywrangler_2020}. Using a data set reflecting a random 10\% of Twitter since 2008 (presently over 150 billion tweets), Storywrangler tracks the prevalence of $n$-grams on a daily scale. $n$-grams are portrayed via rank by popularity, and convey the rise/dynamics of President Trump (further depicted in the PoTUSometer)~\cite{dodds_computational_2021}, or the meteoric rise, and continued influence of Justin Bieber (of surprising relevance to this work). Unlike the Google Books $n$-gram Corpus, StoryWrangler is notable in its ability to track phrases in both original tweets and retweets, conveying aspects of popularity through amplification. 


Beyond simple words and phrases, data have been used to track the progression of ideas. For instance, Leskovec et al.'s paper on ``meme-tracking'' tracked the progression and mutation of popular sayings as they proliferated through news reporting and blogging \cite{leskovec_meme-tracking_2009}.

\begin{figure*}
\centering
\includegraphics[width=\textwidth]{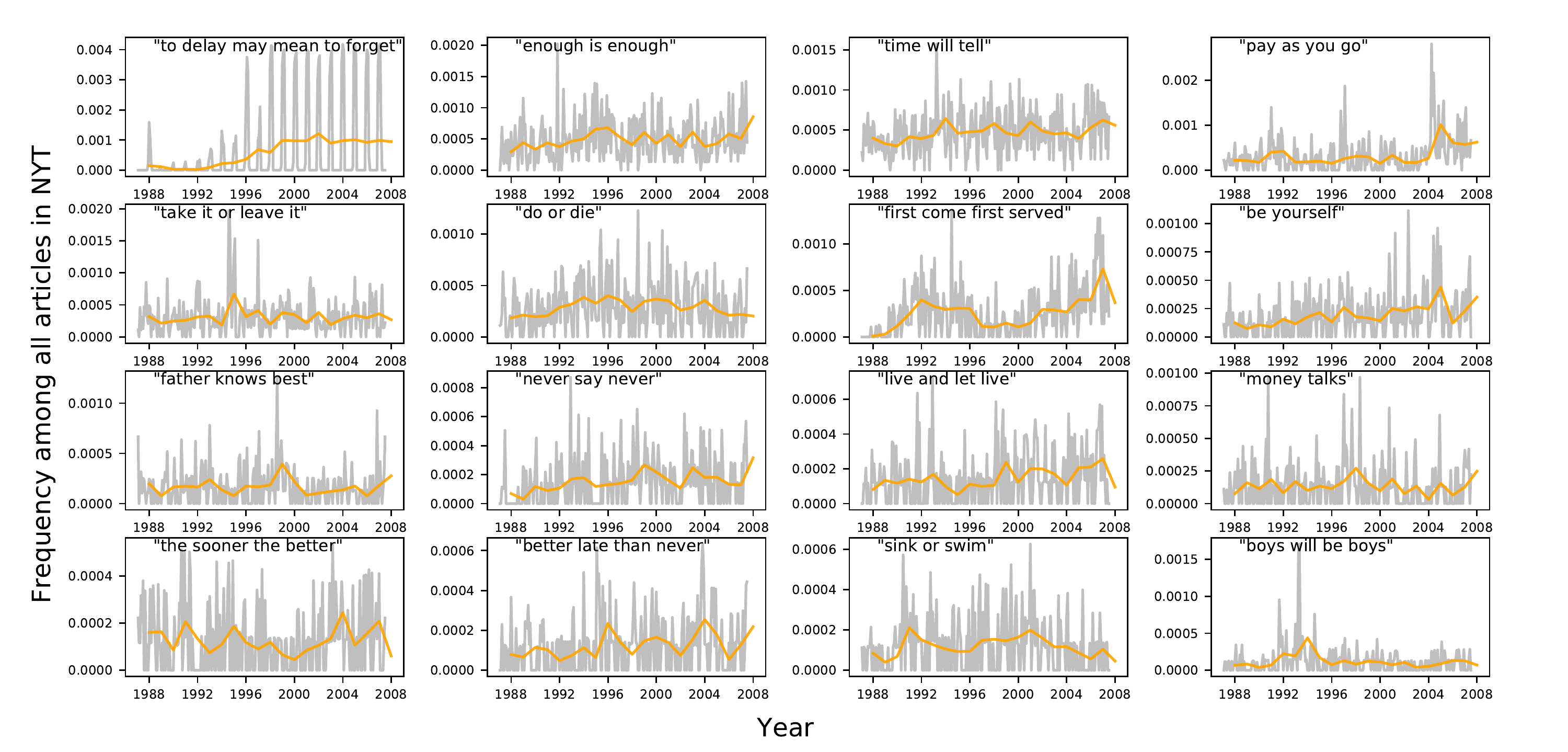}
\caption{\textbf{Time series plots for the 16 most popular proverbs in \textit{The New York Times} from 1997-2007 (ranked by overall count).} The gray represent the data binned by month, and the orange represent the data binned by year. The proverb ``to delay may mean to forget'' owes its yearly rhythm to its role as the NYT's charity tagline. The frequencies are normalized by article count (obits, and non-body included). Plots are ordered in the grid by rank first left to right, then top to bottom.}
\end{figure*}

Recently, ``Computational Folkloristics'' has gained recognition as an area of study, with a 2016 issue of the \textit{Journal of American Folklore} being devoted to the subject \cite{tangherlini_big_2016}. Using classification, networks, geographical data, temporal data, and digitized text, folklorists and other interested academics have explored new possibilities in understanding texts and cultural history. The Danish Folklore Nexus developed by Abello et al.\ provides tools for large-scale analysis of Danish folk tales and stories, aiding in classification of stories, or mapping their similarity to others through networks. Tools like this can augment traditional methods of studying folklore, using data-driven methodology to guide future avenues of folklore research \cite{abello_computational_2012}. This represents a paradigmatic example of a computational tool participating in the continued discourse around folklore, without being an end in and of itself. 

\section{Data and Methods}
\label{sec:methods}

In an effort to quantify the ecology of proverbial language, a list of over 14,000 proverbs was obtained from Mieder's \textit{Dictionary of American Proverbs} \cite{mieder_dictionary_1992}.  Proverbs were stored in an SQL database for ease of access, and matched for frequency with four distinct corpora:\begin{itemize}
\item	The Gutenberg Corpus (English)
\item	The \textit{The New York Times} (1988-2007)
\item   The Google Books $n$-gram Corpus (1800-2000)
\item	Twitter (2008-2020)
\end{itemize}

Individual corpora were collected as follows.


\subsection{Gutenberg}

The Gutenberg corpus comprises over 60,000 collected published documents spanning several centuries. The present study restricts its use to the subset of documents in English. As the metadata for the Gutenberg corpus does not consistently encode the date of original publication, temporal data was collected using author birth dates (gathered from the Gutenberg library for R) \cite{rgut_robin}. These were used in place of publication dates, as the publication dates in the corpus seldom represent the original publication, instead they represent the digitized edition. For temporal analysis, documents without authors and their birth dates were omitted.

\begin{table*}[ht!]
  \setlength{\tabcolsep}{0.5em}
  \renewcommand{\arraystretch}{1}
  \rowcolors{2}{white}{gray!15}
  \begin{tabular}{llr}

\hline
{} &                                                                                                                                                                                                                                                                                                             book &  btwn centrality \\
\hline
1  &  Dictionary of Quotations  &  0.043048 \\
2  &  Familiar Quotations  &  0.022821 \\
3  &  Dictionary of English Proverbs and Proverbial Phrases  &  0.014274 \\
4  &  A Polyglot of Foreign Proverbs &  0.013061 \\
5  &  The Entire Project Gutenberg Works of Mark Twain &  0.013041 \\
6  &  French Idioms and Proverbs &  0.010083 \\
7  &  Roget's Thesaurus &  0.009785 \\
8  &  Webster's Unabridged Dictionary &  0.007978 \\
9  &  U.S. Copyright Renewals 1950 - 1977 &  0.006709 \\
10 &  The Project Gutenberg Complete Works of Gilbert Parker &  0.006278 \\
11 &  Proverb Lore &  0.006028 \\
12 &  Complete Project Gutenberg John Galsworthy Works &  0.003897 \\
13 &  Complete Project Gutenberg Works of George Meredith &  0.003660 \\
14 &  Ulysses &  0.003184 \\
15 &  The Historical Romances of Georg Ebers &  0.003168 \\
16 &  Familiar Quotations &  0.003007 \\
17 &  The Circle of Knowledge &  0.002886 \\
18 &  The Complete Poetic and Dramatic Works of Robert Browning &  0.002749 \\
19 &  Complete Project Gutenberg Oliver Wendell Holmes, Sr. Works &  0.002657 \\
20 &  Motion Pictures, 1960-1969: Catalog of Copyright Entries &  0.002578 \\
\hline

  \end{tabular}
  \caption{\textbf{The 20 most central books by betweenness centrality, from a network of books connected by shared proverbs in Gutenberg.} Notably, James Joyce's \textit{Ulysses} appears alongside several proverb and quotations collections, and the collected works of Mark Twain. }
\end{table*}

The Gutenberg corpus comes with several caveats. Firstly, works were curated by perceived importance. Works also disproportionately represent the 18th and 19th centuries, and for this reason much of our work with Gutenberg focuses on this period. Several authors have much of their extensive oeuvre represented in the corpus (e.g., Anthony Trollope, Mark Twain), which could compromise a more objective view of English writing tendencies of the period.

\subsection{The New York Times}

Data from the \textit{New York Times} were gathered from the \textit{New York Times} Annotated Corpus of 1.8 million articles from 1987-2007 \cite{sandhaus_evan_new_nodate}. The data are organized in NTIF (News Industry Text Format) formatted XML-readable documents. The corpus includes obituaries and other short pieces in addition to more traditional news articles.

\subsection{Google}

The 2020 English Fiction Google $n$-grams corpus consists of every $n$-gram that appears at least 40 times in its set of millions of digitized books. For each $n$-gram the corpus provides on each year it appears in the data set, the frequency with which it appeared that year, and the number of documents it appeared in that year \cite{michel_quantitative_2011}. 
 
\subsection{Twitter}

Data from Twitter was accessed through the Vermont Complex Systems Center's StoryWrangler API\cite{alshaabi_storywrangler_2020}. StoryWrangler receives a randomly selected 1/10th of each day's tweets from Twitter's Decahose API (including retweets), and organizes $n$-grams by rank and frequency. Data for 2-gram and 3-gram proverbs were obtained though the tool, and were aggregated so the collection was case insensitive.

\subsection{Data Processing and Visualization}

The data from all four corpora were processed using Python, and the libraries pandas and matplotlib were used for organization and visualization respectively \cite{reback2020pandas,Hunter:2007}. 
In our processing of Gutenberg and the \textit{New York Times}, punctuation in both proverbs and texts was removed. Twitter data were punctuation insensitive. Regular expressions were used to capture variations in punctuation when processing the Google Books $n$-gram Corpus. 

Where relative frequency is used, it is calculated as: $$f_{rel} = f_t/n_t$$ which is the frequency $f$ for time period $t$ divided by the number of documents $n$ found during time period $t$. Zipf distributions were plotted using ranks of proverbs in a corpus, with rank 1 being the most frequent, as well as their frequency. Zipf distribution plots are shown on a log-log scale as is standard. 

Networks of books and proverbs, as well as authors and proverbs, were made using books/authors as nodes, connected by proverbs they have in common. The networks are unweighted, and do not reflect instances where books/authors share multiple proverbs. Betweenness centrality in these networks is calculated as $$b(v) = \sum_{s\neq v \neq t}{\frac{\sigma_{st}(v)}{\sigma_{st}}},$$ or the proportion of shortest paths between any two other nodes in the network that pass through a given node.

Most processing was performed using the Vermont Advanced Computing Core (VACC) located at the University of Vermont. 

\begin{figure*}[tp!]

\includegraphics[width=\textwidth]{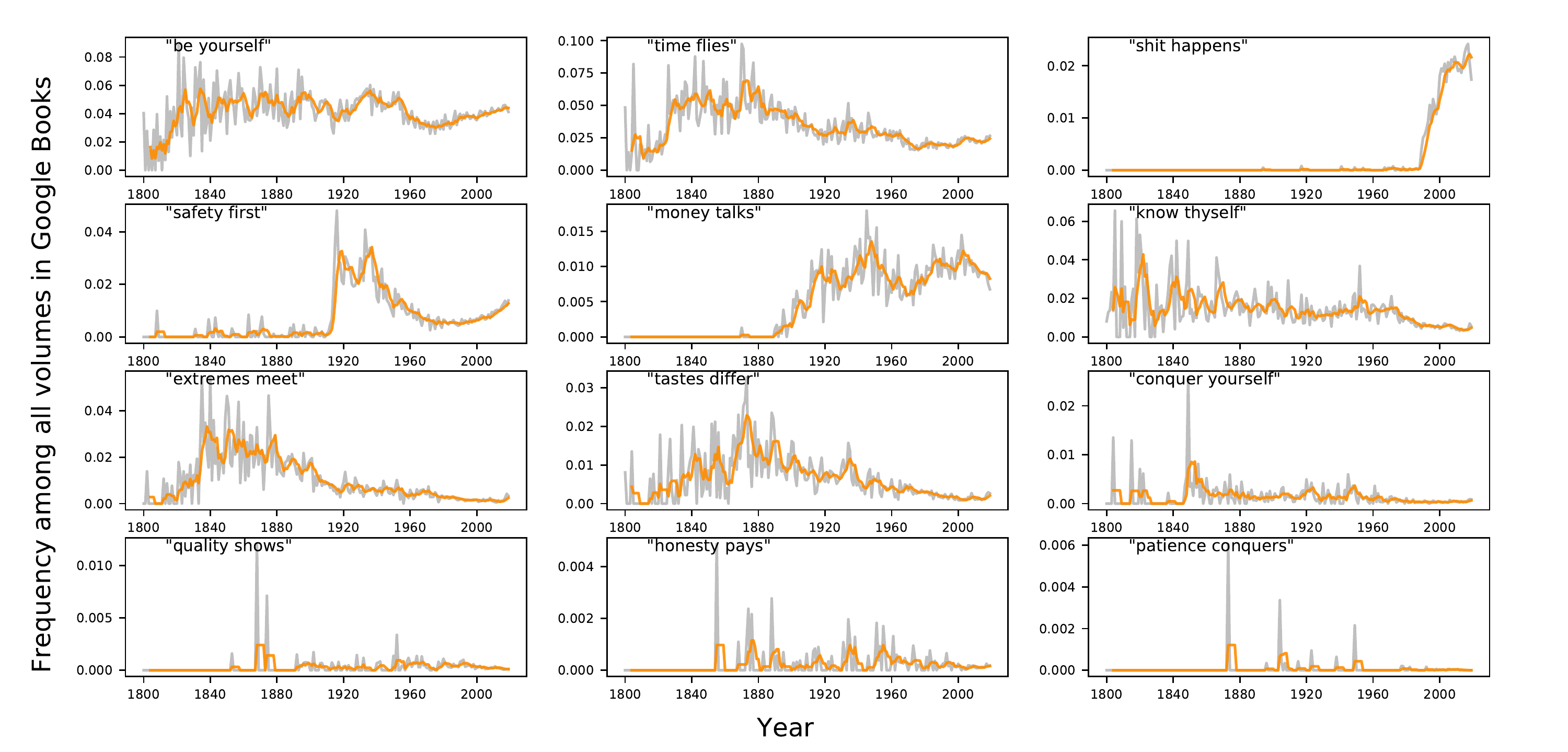}
\caption{\textbf{Time series plots for the 12 most popular 2-gram proverbs in the Google Books $n$-gram Corpus (ranked by overall count)}. The gray represent the yearly frequency, while the orange represent the five-year rolling average. The dramatic increase in use of the proverbs ``shit happens'' and ``safety first'' correspond with previous scholarship on their emergence. Plots are ordered in the grid by rank first left to right, then top to bottom.}
\end{figure*}

\section{Results}
\label{sec:results}

\subsection{Zipf distributions}

Fig.~\ref{fig:paremiology.zipf}
shows Zipf distributions for entries from Mieder's \textit{Dictionary of American Proverbs} using 3-gram anchors for each of the four corpora studied
While the distributions exhibit heavy tails,
we do not observe robust power-law scaling over many orders of magnitude.
We find the largest number of 
distinct proverbs appearing in Gutenberg and the New York Times, on the order of thousands,
with the Google Books and Twitter examples showing roughly an order of magnitude fewer.
We note that Zipf's law for words does not itself 
extend over many orders of magnitude~\cite{williams2015b}, typically only 2 or 3,
and that it is meaningful, mixed length phrases that present many orders of magnitude of scaling~\cite{ryland_williams_zipfs_2015}.
The Zipf distributions for proverbs are thus comparable to what we see for single words.

With a more sophisticated proverb detection, one that captures minor variations in phrase structure, we would expect to see some adjustments to the Zipf distributions we have observed, though a priori it is not clear how. 
Short, robust proverbs (``time flies'') will be well counted, while longer ones for which, say, constituent function words might be changed based on context or era (``he/she/they who hesitates \ldots'') would only see their
apparent observed frequency of usage grow.

\subsection{Gutenberg}

While the most popular entry in the Gutenberg corpus and the Google Books $n$-gram corpus was the phrase ``hold your tongue'', this phrase is classified as a proverbial expression rather than a proverb (its use requires outside context). For clarity of focus the phrase has been excluded from figures in this section
. ``Sink or swim'', another proverbial expression, has been left in. In light of the limitations of the Gutenberg corpus detailed in Methods, it is difficult to make claims about the trends of proverb use over time (Figure 3). However, it is clear from the data shown in Figure 3 that proverbs appear in a remarkable portion of the documents in the corpus. ``The sooner the better'' for example, appears in nearly one in every ten documents in the early 1800s.

The data for proverbs in the Gutenberg corpus were used to construct a network with documents as nodes, connected if a given proverb appears in both documents. When betweenness centrality was calculated for nodes in the network, surprisingly James Joyce's \textit{Ulysses} had the 14th highest centrality, close to several dictionaries of proverbs and quotations, and the collected works of Mark Twain (Table 1). Creasy \cite{creasy_vary_2008} documented Joyce's use of proverbs in \textit{Ulysses} from a critical perspective, noting that they are often altered, and blend high and low culture in the work. As Joyce uses many fewer proverbs than a comprehensive proverbial dictionary, the book's centrality in this network implies that Joyce's use of proverbs is far from arbitrary, and that his choice of proverbs is purposefully situated in the broader context of English proverbial knowledge.

\begin{figure*}
\centering
\includegraphics[width=\textwidth]{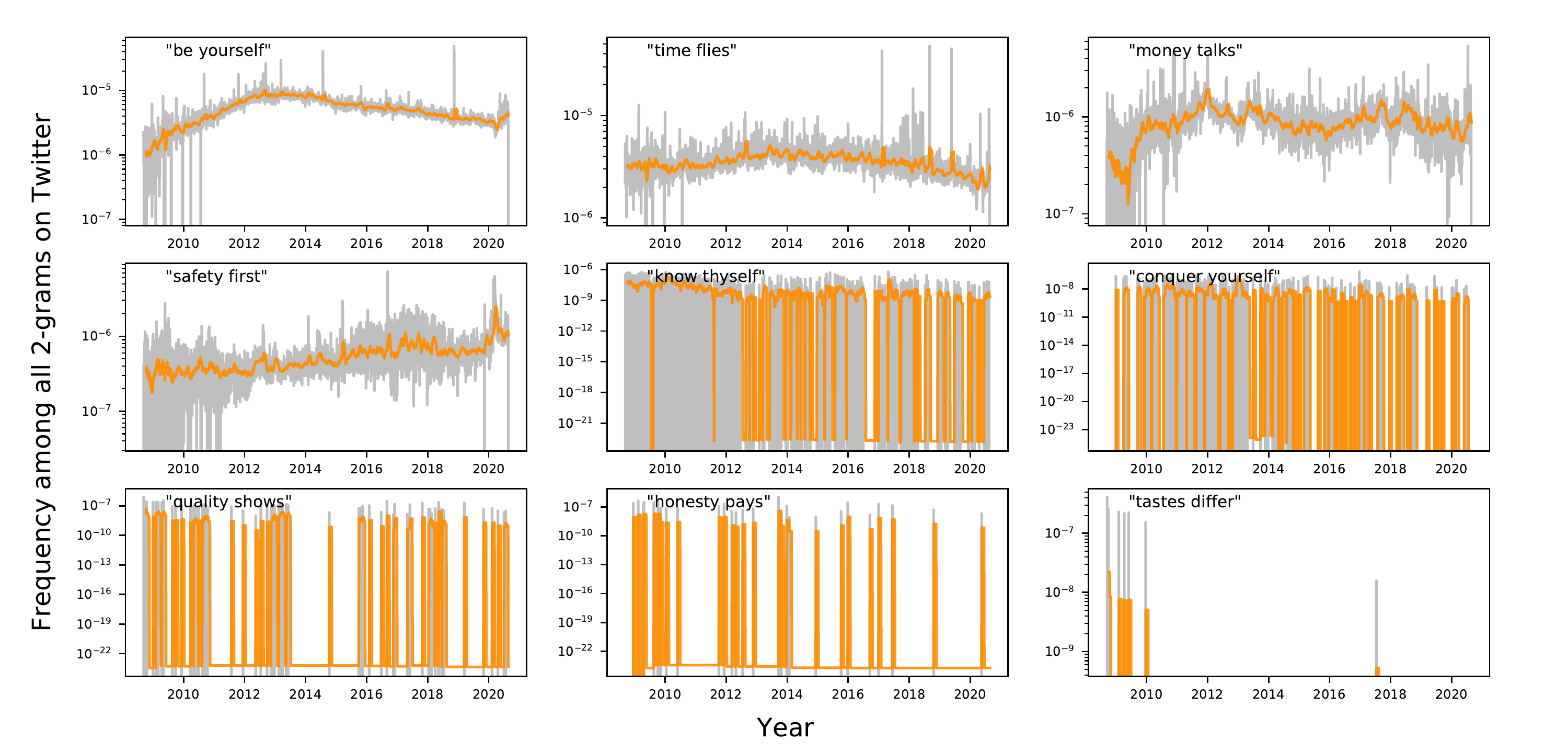}
\caption{\textbf{Time series plots for the nine most popular 2-gram proverbs on Twitter (ranked by overall count).} The gray represents the daily frequency, while the orange represents the 30 day rolling average. The proverbs ``be yourself'' and ``time flies'' maintain popularity over the period studied. Notably, the ``safety first'' shows an increase in popularity in early 2020, possibly relating to the coronavirus pandemic. Plots are ordered in the grid by rank first left to right, then top to bottom.}
\end{figure*}

\subsection{\textit{The New York Times}}

Figure 4 shows time series plots for the 16 most common proverbs in \textit{New York Times} Annotated Corpus. Shown are frequency binned by month and year, and normalized by article count. All articles are included in the count including smaller articles like obituaries (the average article count is 248 per issue). It is by no means a surprise that proverbs appear frequently in journalism; in fact Lau's study found as much \cite{lau_its_1996}. Not present in that work, however is a temporal dimension (not to mention a different time period). It is clear in Figure 4 that the proverbs represented are used on a monthly or semi-monthly basis, and are rarely if ever absent in a year's publications. In these representations of proverb use, it is easier to identify use patterns and perhaps to extract narratives from their dynamics. The easiest, if somewhat trivial case is ``to delay may mean to forget'' owes its yearly rhythm to its role as the NYT's charity tagline. Its frequency of use increased markedly over the period studied, though stayed confined to the winter holiday months. 

With the exception of ``to delay may mean to forget'', and consistent with accepted definitions of the proverb, the consistency with which proverbs are used in the \textit{New York Times} suggests they are employed widely for their utility in mapping general wisdom to a specific context. Nonetheless, prominent spikes in frequency can be associated with historical events. For instance the brief several-fold increase in the use of ``boys will be boys'' around November of 1992 is likely attributed to a contentious and widely publicized sexual assault case at the time, which prompted additional discussion of rape culture \cite{glaberson_assault_1992, hanley_jury_1992}. 

The maximum in use of ``pay as you go'' seems to correspond with concurrent discussion of a local gas tax levy in New Jersey, and national discussion of President Bush's second term proposed tax cuts. Its increase in use in 1996 seems to owe to discussion of the Environmental Bond Act being proposed in New York at the time \cite{noauthor_vote_1996, henry_how_1996}. 

\subsection{Google}

In Figure 5 are time series plots for the 12 most common 2-gram proverbs in the Google $n$-grams corpus. Here the gray represents yearly frequency (counted once per volume), and the orange represents the five-year rolling average, normalized by the number of volumes in a given year. One can see clearly from the figure the emergence of several more recent proverbs: ``safety first'', ``money talks'', and ``shit happens''. 

``Safety first'' exhibits a precipitous rise in usage in the early 20th century. Specifically, in 1912, the National Safety Council (NSC) in the US adopted the phrase as its slogan to promote standards of worker safety, though the Safety First Movement was initiated by US Steel in 1906. Its origin has been traced back to at least 1818 \cite{mieder_right_2019}. The data shown in Figure 5 support the history of its popularization \cite{swuste_safety_2010, mieder_dictionary_1992}.

Previous scholarship on the proverb ``shit happens'' traced its origin to the year 1944, and its rise in popularity corresponds to its humorous use as a bumper sticker, and cultural controversy (and legal battles) associated with it \cite{mieder_proverbs_2004, cunningham_v_state_1991}. It also famously appeared in the movie \textit{Forrest Gump} \cite{zemeckis_forrest_1994}.

Figure 6 shows time series plots for the 16 most popular 3-gram proverbs in the Google Books $n$-gram corpus. Though the proverb ``never say never'' originated in 1887 \cite{mieder_dictionary_1992}, it is evident that it gained far wider popularity in the late 1900s. Though the proverb ``enough is enough'' dates at least to 1546 \cite{mieder_dictionary_1992}, its popularity seems to vastly increase throughout the 20th century. The proverb ``divide and conquer'' seems to have briefly gained popularity around the World War II era. 

\subsection{Twitter}

\begin{figure*}
\includegraphics[width=\textwidth]{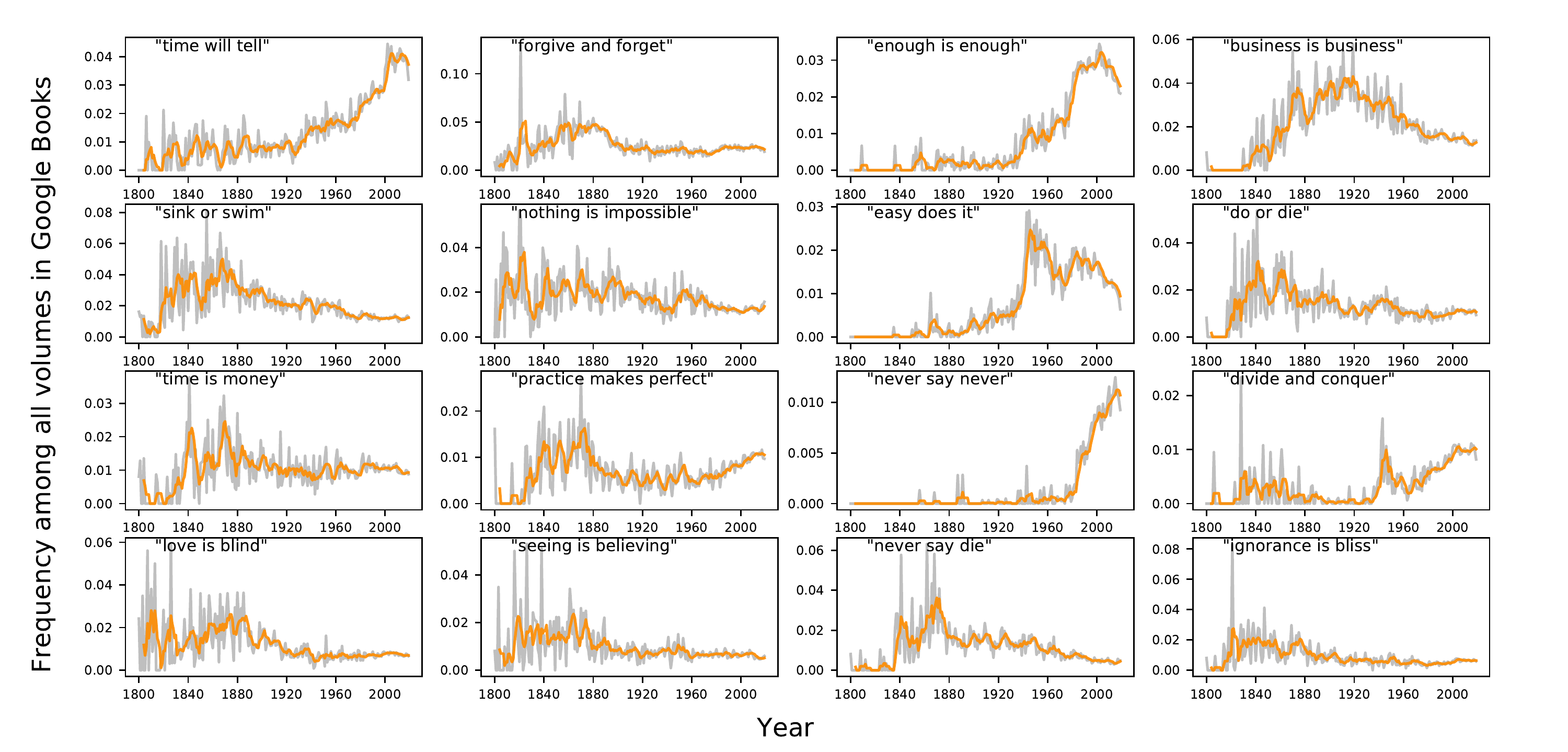}
\caption{\textbf{Time series plots for the 16 most popular 3-gram proverbs in the Google Books Ngram Corpus (ranked by overall count)} The gray represents the yearly frequency, while the orange represents the 5 year rolling average. The rise in popularity of the proverb ``never say never'' is shown. A period of increased usage of the proverb ``divide and conquer'' corresponds with the World War II era. Plots are ordered in the grid by rank first left to right, then top to bottom.}
\end{figure*}

On Twitter, the four most common 2-gram proverbs, on average, don't seem to exhibit much variability in their usage (Figure 7). The proverbs ``be yourself'' and ``time flies'' seem to remain above $10^{-6}$, or 1 in every million 2-grams on Twitter during the period studied. An increase in usage of ``safety first'' in early 2020 may be related to the onset of the coronavirus pandemic during the same period. 

Exhibited on Twitter (Figure 8), the convenience of proverbs as succinct narratives has made them useful in several titular media events in the past decade. Of note, Figure 8 shows marked shifts in frequency of ``never say never'', and ``love is blind''. ``Never say never'' owes its initial attention in 2010 to Justin Bieber's single of the same title (\textit{Justin Bieber: Never Say Never}), repeated as his slogan and title of a biographical documentary. This was not the first film to utilize the proverb in its title; Sean Connery's final performance as James Bond was titled \textit{Never Say Never Again} (1983).  

Figure 9 shows the dynamics of ``never say never'' on Twitter in more detail. We observe first its meteoric rise in popularity at the time of \textit{Never Say Never}'s (song) release as the lead single off the soundtrack for a modern remake of the \textit{Karate Kid} movie (roughly two magnitudes in a single day). At the time of the single's official release on June $8^{th}$, 2010, ``never say never'' was the $63^{rd}$ most used 3-gram on Twitter. When \textit{Justin Bieber: Never Say Never} was released on January 31, 2011, ``never say never'' was the $34^{th}$ most common 3-gram on Twitter; for comparison, ``I love you'' was $22^{nd}$ at the time. 

Remarkably, the popularity of ``never say never'' on Twitter decayed so slowly that it did not reach its pre-Bieber frequency until 2016. The continued presence of the proverb in Twitter discourse suggests that in the wake of its initial rise, it was more frequently adopted to general non-Bieber usage. (A similarly popular 3-gram, non-proverbial song of that year ``rock that body'' appeared and disappeared from the Twitter discourse in the span of a few months). While the enormity and fervor of Bieber's fanbase at the time (a period called ``Bieber fever''\cite{Tweedle2012AMM}) certainly contributed to its popularity, its continued use over a five-year period is compelling evidence that the proverb became a more integral part of the Twitter lexicon for a time. 

\begin{figure*}
\centering
\includegraphics[width=\textwidth]{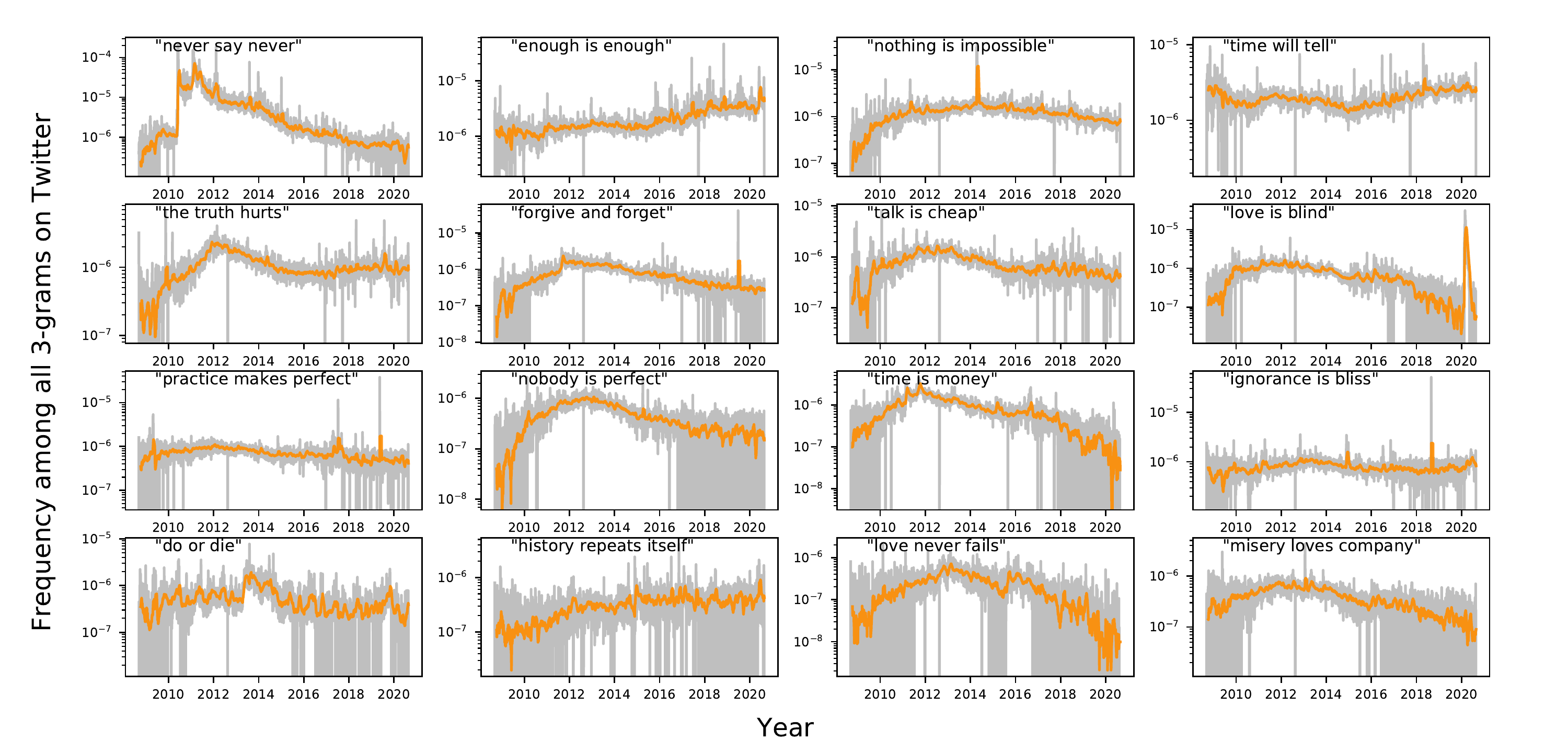}
\caption{\textbf{Time series plots for 3-gram proverbs on Twitter (ranked by overall count).} The gray represents the daily frequency, while the orange represents the 30 day rolling average. The proverb ``never say never'' owes its meteoric rise in popularity in 2010 to popular musician Justin Bieber's single and biographical documentary of the same name. ``never say never'' remains the most popular proverb on Twitter until 2016, when it is supplanted by ``enough is enough'' which has steadily gained popularity in the last decade, owed in part to its constant use by Senator Bernie Sanders, and punctuated by reactions to tragedies related to gun and police violence. Plots are ordered in the grid by rank first left to right, then top to bottom.}
\end{figure*}

In 2020, ``Love is Blind'' became the title of a literally minded reality dating show in which participants were quarantined in private rooms, only communicating via audio interfaces \cite{coelen_love_2020}. In this instance, the proverb was not only an apt description of the show's narrative, but a template for its formation. Additionally, it came to represent a narrative solution to the isolation imposed by the concurrent pandemic. However, the increase in the phrase's popularity seems only to have lasted for the month of the show's release, after which it seems to settle at its former rate of use. The proverb itself is ancient, and translations exist nearly every European language.

While with ``never say never'' (the most popular proverb on Twitter), we see a sudden rise and slow decay, we see a different pattern in the second most popular proverb, ``enough is enough''. From 2016 to the present, we see a steady increase in the frequency of ``enough is enough'' on Twitter. Recent work by Mieder attributes its renewed popularity in part to its constant use by Bernie Sanders \cite{mieder_right_2019}. Unlike ``never say never'' there does not seem to be a single event that precipitates this trend. However, an investigation into the several local maxima suggest a possible narrative correspondence. Many of these local maxima correspond to events related to either police violence or mass shootings.

Famously, survivors of the Parkland shooting in 2018 appeared on the cover of \textit{Time} magazine with a simple title: ``Enough.'' \cite{alter_young_2018}. Coverage of the March for Our Lives against gun violence in the \textit{New York Times} included the title: \textit{March for Our Lives Highlights: Students Protesting Guns Say ``Enough Is Enough''} \cite{ourlives_march_2018}. When protesters marched in DC in the wake of the murder of George Floyd, Politico's coverage was titled: \textit{`Enough is enough': Thousands descend on D.C. for largest George Floyd protest yet} \cite{semones_enough_202} . Inasmuch as proverbs can create metaphorical mappings from a paradigmatic situation (or narrative) onto a present one, ``enough is enough'' represents a compelling narrative of continued injustice, and a critical point of retaliation. However, the data from Twitter display a narrative of repeated tragedy in spite of public outcry. The proverb was most popular during the 2018 US midterm elections.

\section{Concluding remarks}
\label{sec:concludingremarks}

This study is by no means the first exploring the potential of new and growing digital databases for the future of phraseology. In fact, in one of the most recent textbooks on parameiology, there is a section on proverbs and corpus linguistics. However, as yet, we believe there has been no large-scale effort to examine the dynamics of proverb use across corpora of several domains. Pioneering work by \v{C}erm\'{a}k and Moon validated the usage computational resources to augment quantitative efforts. That work was limited by the available computational and digital resources.

\begin{figure*}
\centering
\includegraphics[width=\textwidth]{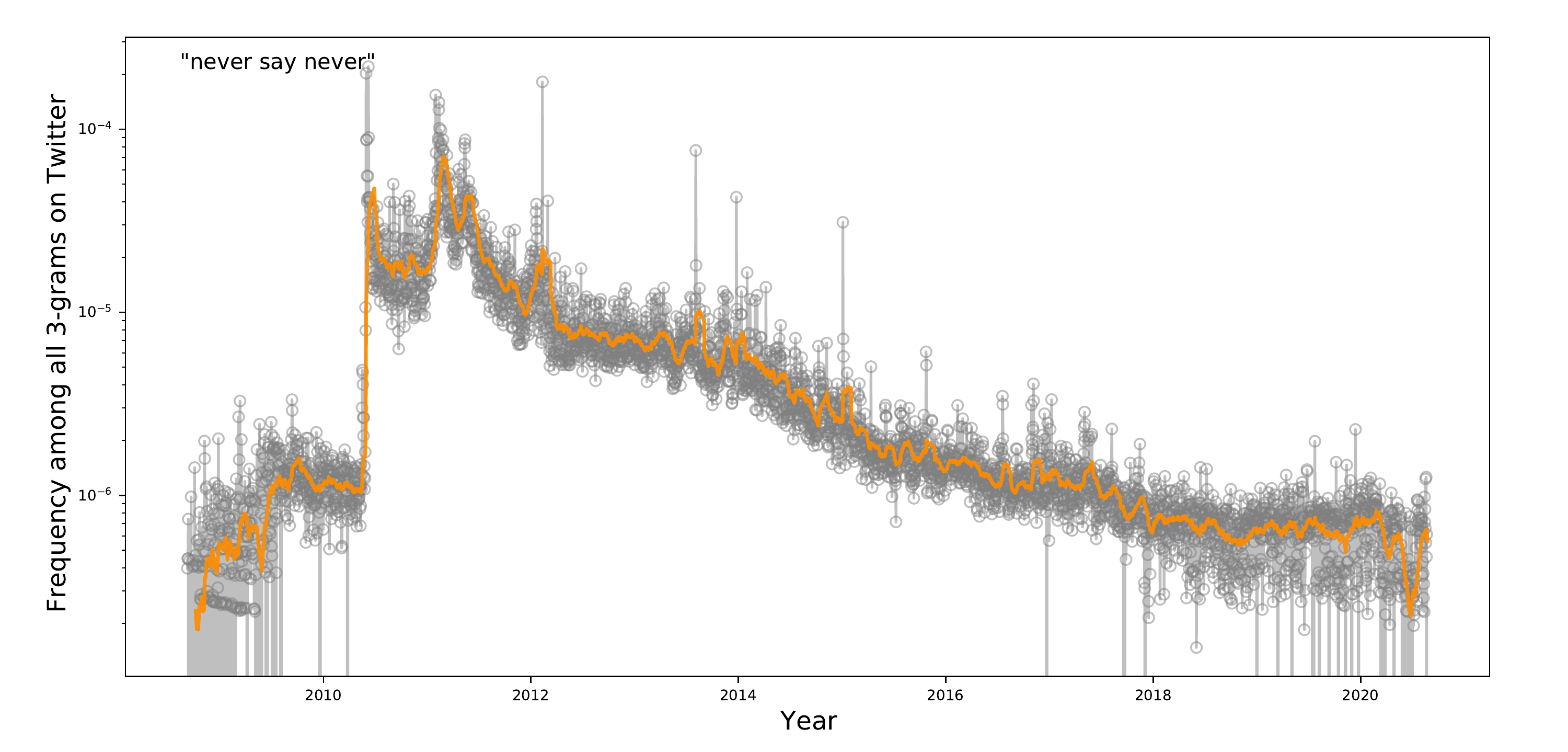}
\caption{\textbf{Daily relative frequency of the 3-gram ``never say never'' on Twitter}. While ``never say never'' was already popular on Twitter as of 2008, its popularity was amplified in 2010 by the release of Justin Bieber's single entitled ``Never say never'', and his subsequent biographical documentary of the same name. Remarkably, it remained the most popular proverb on Twitter for almost six years, punctuated by anniversaries and reruns of the movie, until it was surpassed by ``enough is enough'' in 2016. }
\end{figure*}

Much attention has been paid to the use of words and $n$-grams in general in large corpora, but it is difficult to extract from them instances of individual narrative or metaphorical language use. Proverbs, in their tendency to act as both narrative and metaphor, and in their often relatively fixed structure, are perhaps an ideal test case for our ability to observe broader cultural narratives through the piecemeal, routine stories employed by the folk.

Through novel or context-specific words and phrases, we are able to observe discourse around specific phenomena (``pizzagate’‘, ``pandemic’‘, or ``Make America Great Again’’). In contrast, through proverbs we may be able to observe how we organize specific phenomena into the paradigmatic narratives represented by proverbs.

Much of proverb scholarship has been concerned with the idea of a ``paremiological minimum'': 
A minimum proverbial lexicon for a language and culture. 
Certainly, as shown by Lau~\cite{lau_cheaters_2004}, and again in the present study, computational studies of the frequency of proverb use can contribute to the understanding of these minima, as those proverbs which seem ubiquitous in large corpora ought to be understood by speakers of a language. 
Furthermore, temporal analysis of their frequency may further validate that their frequency is related to enduring currency among the folk, rather than correspondence with a specific occurrence. 
Another concern in paremiology and phraeseology is the origins of sayings. Work like the present study can serve to both validate and expand on previous scholarship on the history of phrases. 

In the study of the statistical distribution of natural language, there exists the idea of a kernel lexicon, a subset of words that are essential to communication using a given language. Much literature on the study of culture and education has focused on what one might consider a ``minimum of cultural literacy''. Special attention has been paid to which proverbs constitute part of that minimum. 
It is clear from this study that the most common proverbs vary considerably between corpora. 
However, given the prevalence of these popular proverbs in their respective contexts, we can posit that English learners would benefit in their comprehension of the language if they were familiar with these proverbs.

A natural limitation of this study, and indeed any study that uses extant data to study language, is the issue of representativeness. In this study that limitation is twofold: 
Both the lexicon for directing the search, and the data being searched are inherently limited. While \textit{The Dictionary of American Proverbs} is extensive, and represents much that is known of proverbs in America, it naturally excludes new proverbs and does not account for many ways in which the structure of the proverbs it contains may be manipulated in their practical use. There are however lexical resources that address recent proverbs, for example \textit{The Dictionary of Modern Proverbs}, and the methodology of this study may be readily applied to such lexica \cite{doyle_dictionary_2012}. 
Previous studies on proverb frequency have relied on composite corpora, namely variations of the BNC (British National Corpus), which contains manually curated selections from several domains of text. 
The present approach of studying data from distinct domains allows for both a more limited and more useful interpretation of the results: 
We can only claim that results are representative of proverb use on Twitter for instance, rather than proverb use in English as a whole---an impossible achievement. 
Certainly, fieldwork (digital and otherwise) continues to be important in identifying new proverbs and changing structures of existing proverbs. 
This task may be aided in the future by tools like StoryWrangler, that track $n$-gram rank, likely capturing new proverbs in the process. The task then would be extracting likely proverbs from these data, which would require both linguistic, cultural, and computational expertise. 

Analyses of the frequency and rank of proverbs in this study verify that with ever increasing amounts of machine-readable textual data, we may produce longitudinal phraseological studies. 

Furthermore, as machine/robot comprehension of natural language becomes increasingly important, this area too, would benefit from an expanded lexicon that includes proverbs and routine formulae, and understanding of metaphor may be assisted by a more basic understanding of the mapping from general to specific situations that exists in the use of proverbs. 

\acknowledgments
Thank you to David Dewhurst, Josh Minot, Michael Arnold, Nicholas Allgaier, Thayer Alshaabi for providing invaluable guidance in the writing of this paper. The authors are grateful for the computing resources provided by the
Vermont Advanced Computing Core
which was supported in part by NSF award No. OAC-1827314,
and financial support from the
Massachusetts Mutual Life Insurance Company to CMD and PSD.

\clearpage

\pagenumbering{arabic}
\setcounter{figure}{0}
\setcounter{table}{0}
\renewcommand{\thepage}{S\arabic{page}}
\renewcommand{\thesection}{S\arabic{section}}
\renewcommand{\thetable}{S\Roman{table}}
\renewcommand{\thefigure}{S\arabic{figure}}

\section*{Supplementary Material}

Tables SI-SIV show the total count of the 50 most popular proverbs in
their respective corpora.

Supplementary Figures S1-S4 are in the same format as the time series
plots in the body, but show data for proverbs ranked 17-32 in their
respective corpora.

\newpage

\begin{table}[h!]
  \setlength{\tabcolsep}{0.5em}
  \renewcommand{\arraystretch}{1}
  \rowcolors{2}{white}{gray!15}
\caption{\textbf{The top 50 proverbs and proverbial expressions (from the Dictionary of American Proverbs) in the entire Gutenberg Corpus.}}
  \begin{tabular}{llr}

\hline
{} &                            Proverb &  Count \\
\hline
1  &                   hold your tongue &  2,284 \\
2  &              the sooner the better &  1,536 \\
3  &                        be yourself &    739 \\
4  &             let bygones be bygones &    685 \\
5  &                         time flies &    603 \\
6  &           alls well that ends well &    588 \\
7  &                one thing at a time &    580 \\
8  &               business is business &    534 \\
9  &                       sink or swim &    531 \\
10 &                 forgive and forget &    477 \\
11 &                take it or leave it &    436 \\
12 &              nothing is impossible &    421 \\
13 &             better late than never &    419 \\
14 &              every man for himself &    414 \\
15 &                       know thyself &    394 \\
16 &              share and share alike &    372 \\
17 &                      slow but sure &    363 \\
18 &                  live and let live &    356 \\
19 &               the more the merrier &    352 \\
20 &                    the die is cast &    348 \\
21 &         honesty is the best policy &    339 \\
22 &                 to be or not to be &    335 \\
23 &                          do or die &    322 \\
24 &                      never say die &    319 \\
25 &                      extremes meet &    289 \\
26 &                  art for arts sake &    286 \\
27 &          all men are created equal &    265 \\
28 &              let well enough alone &    260 \\
29 &                      time is money &    250 \\
30 &            no accounting for taste &    249 \\
31 &                 peace at any price &    244 \\
32 &                      tastes differ &    241 \\
33 &             history repeats itself &    235 \\
34 &                  boys will be boys &    235 \\
35 &             charity begins at home &    231 \\
36 &                      love is blind &    228 \\
37 &        the end justifies the means &    227 \\
38 &     one good turn deserves another &    224 \\
39 &        blood is thicker than water &    221 \\
40 &            not wisely but too well &    219 \\
41 &  all things work together for good &    213 \\
42 &            first come first served &    201 \\
43 &        keep the wolf from the door &    196 \\
44 &             dead men tell no tales &    195 \\
45 &          the wages of sin is death &    191 \\
46 &                seeing is believing &    187 \\
47 &             keep a stiff upper lip &    186 \\
48 &                 ignorance is bliss &    185 \\
49 &   where theres a will theres a way &    183 \\
50 &                    murder will out &    179 \\
\hline
\end{tabular}

\end{table}

\newpage

\begin{table}[h!]
  \setlength{\tabcolsep}{0.5em}
  \renewcommand{\arraystretch}{1}
  \rowcolors{2}{white}{gray!15}
  
\caption{\textbf{The top 50 proverbs and proverbial expressions (from the Dictionary of American Proverbs) in The New York Times from 1987-2007}}
  \begin{tabular}{llr}

\hline
{} &                            Proverb &  Count \\
\hline
1  &        to delay may mean to forget &  1,075 \\
2  &                   enough is enough &    891 \\
3  &                     time will tell &    864 \\
4  &                      pay as you go &    597 \\
5  &                take it or leave it &    565 \\
6  &                          do or die &    528 \\
7  &            first come first served &    463 \\
8  &                        be yourself &    348 \\
9  &                  father knows best &    307 \\
10 &                    never say never &    276 \\
11 &                  live and let live &    272 \\
12 &                        money talks &    244 \\
13 &              the sooner the better &    240 \\
14 &             better late than never &    224 \\
15 &                       sink or swim &    218 \\
16 &                  boys will be boys &    213 \\
17 &                         time flies &    205 \\
18 &             time is of the essence &    204 \\
19 &                 divide and conquer &    198 \\
20 &           gentlemen prefer blondes &    192 \\
21 &                 to be or not to be &    187 \\
22 &                the show must go on &    185 \\
23 &                      time is money &    174 \\
24 &                      talk is cheap &    167 \\
25 &              every man for himself &    166 \\
26 &            leave well enough alone &    163 \\
27 &                  put up or shut up &    161 \\
28 &               business is business &    159 \\
29 &            accentuate the positive &    157 \\
30 &                 forgive and forget &    151 \\
31 &           you get what you pay for &    142 \\
32 &                       safety first &    142 \\
33 &            too little and too late &    140 \\
34 &               there is no easy way &    132 \\
35 &  let the chips fall where they may &    131 \\
36 &          all men are created equal &    129 \\
37 &               the more the merrier &    128 \\
38 &             history repeats itself &    122 \\
39 &             let bygones be bygones &    117 \\
40 &                one thing at a time &    113 \\
41 &         let nature take its course &    106 \\
42 &                      never say die &    106 \\
43 &                seeing is believing &    102 \\
44 &              nothing is impossible &    100 \\
45 &                        war is hell &     95 \\
46 &           the worst is yet to come &     85 \\
47 &    actions speak louder than words &     82 \\
48 &             gone but not forgotten &     82 \\
49 &                    to each his own &     80 \\
50 &               let the buyer beware &     80 \\
\hline
\end{tabular}

\end{table}

\newpage

\begin{table}[h!]
  \setlength{\tabcolsep}{0.5em}
  \renewcommand{\arraystretch}{1}
  \rowcolors{2}{white}{gray!15}
  \caption{\textbf{The top 50 3-gram proverbs and proverbial expressions (from the Dictionary of American Proverbs) in the Google Books Ngram Corpus.}}

  \begin{tabular}{llr}

\hline
 &                      Proverb &    Count \\
\hline
1  &             hold your tongue &  131,426 \\
2  &               time will tell &   65,640 \\
3  &           forgive and forget &   45,189 \\
4  &             enough is enough &   43,149 \\
5  &         business is business &   30,101 \\
6  &                 sink or swim &   26,315 \\
7  &        nothing is impossible &   25,695 \\
8  &                 easy does it &   23,655 \\
9  &                    do or die &   21,672 \\
10 &                time is money &   18,856 \\
11 &       practice makes perfect &   17,469 \\
12 &              never say never &   16,649 \\
13 &           divide and conquer &   15,673 \\
14 &                love is blind &   14,439 \\
15 &          seeing is believing &   12,951 \\
16 &                never say die &   12,329 \\
17 &           ignorance is bliss &   11,838 \\
18 &       history repeats itself &   11,529 \\
19 &                 fair is fair &   10,456 \\
20 &                slow but sure &    9,898 \\
21 &      forewarned is forearmed &    9,860 \\
22 &            love conquers all &    9,839 \\
23 &         misery loves company &    9,654 \\
24 &              facts are facts &    8,944 \\
25 &               time will pass &    8,389 \\
26 &            orders are orders &    7,620 \\
27 &              the truth hurts &    7,292 \\
28 &              blood will tell &    6,840 \\
29 &            father knows best &    6,783 \\
30 &            try anything once &    6,388 \\
31 &              murder will out &    6,349 \\
32 &            silence is golden &    6,278 \\
33 &                  war is hell &    6,136 \\
34 &     business before pleasure &    5,811 \\
35 &                talk is cheap &    5,723 \\
36 &             revenge is sweet &    5,400 \\
37 &  familiarity breeds contempt &    5,095 \\
38 &            might makes right &    4,768 \\
39 &          consider the source &    4,677 \\
40 &                 toe the mark &    4,549 \\
41 &           every little helps &    4,139 \\
42 &              time marches on &    4,019 \\
43 &           nothing is perfect &    4,007 \\
44 &               money is power &    3,757 \\
45 &    circumstances alter cases &    3,668 \\
46 &          respect your elders &    3,644 \\
47 &     gentlemen prefer blondes &    2,922 \\
48 &            mother knows best &    2,908 \\
49 &             love never fails &    2,848 \\
50 &            nobody is perfect &    2,801 \\
\hline
\end{tabular}

\end{table}

\newpage

\begin{table}[h!]
  \setlength{\tabcolsep}{0.5em}
  \renewcommand{\arraystretch}{1}
  \rowcolors{2}{white}{gray!15}
\caption{\textbf{The top 50 proverbs and proverbial expressions (from the Dictionary of American Proverbs) on Twitter from 2008-2021.} }
  \begin{tabular}{llr}

\hline
 {}      &                      Proverb &      Count \\
\hline
1  &              never say never &  2,549,095 \\
2  &             enough is enough &  2,182,460 \\
3  &        nothing is impossible &    978,533 \\
4  &               time will tell &    869,662 \\
5  &              the truth hurts &    748,285 \\
6  &           forgive and forget &    557,294 \\
7  &                talk is cheap &    465,608 \\
8  &                love is blind &    426,010 \\
9  &       practice makes perfect &    405,635 \\
10 &            nobody is perfect &    399,324 \\
11 &                time is money &    383,632 \\
12 &           ignorance is bliss &    377,037 \\
13 &                    do or die &    316,328 \\
14 &       history repeats itself &    307,467 \\
15 &             love never fails &    255,795 \\
16 &         misery loves company &    226,217 \\
17 &           divide and conquer &     94,085 \\
18 &              facts are facts &     90,513 \\
19 &          respect your elders &     89,372 \\
20 &          seeing is believing &     86,169 \\
21 &               time will pass &     84,432 \\
22 &            silence is golden &     82,346 \\
23 &            love conquers all &     80,964 \\
24 &             revenge is sweet &     69,820 \\
25 &             health is wealth &     66,274 \\
26 &                never say die &     65,115 \\
27 &        prayer changes things &     63,757 \\
28 &           iron sharpens iron &     57,065 \\
29 &                 sink or swim &     50,361 \\
30 &         tomorrow never comes &     50,297 \\
31 &         business is business &     39,525 \\
32 &             hold your tongue &     34,344 \\
33 &           nothing is perfect &     34,050 \\
34 &            try anything once &     33,370 \\
35 &            mother knows best &     26,848 \\
36 &           every little helps &     23,672 \\
37 &             never waste time &     22,244 \\
38 &                 fair is fair &     18,125 \\
39 &                slow but sure &     14,404 \\
40 &          consider the source &     14,201 \\
41 &             justice is blind &     11,604 \\
42 &               money is power &     10,186 \\
43 &           time works wonders &     10,079 \\
44 &      time changes everything &      9,512 \\
45 &           like attracts like &      8,320 \\
46 &  familiarity breeds contempt &      8,166 \\
47 &                  war is hell &      7,439 \\
48 &                 easy does it &      6,071 \\
49 &     gentlemen prefer blondes &      5,273 \\
50 &       courtesy costs nothing &      3,890 \\
\hline
\end{tabular}

\end{table}

\newpage
\begin{figure*}[h!]
\centering
\includegraphics[width=\textwidth]{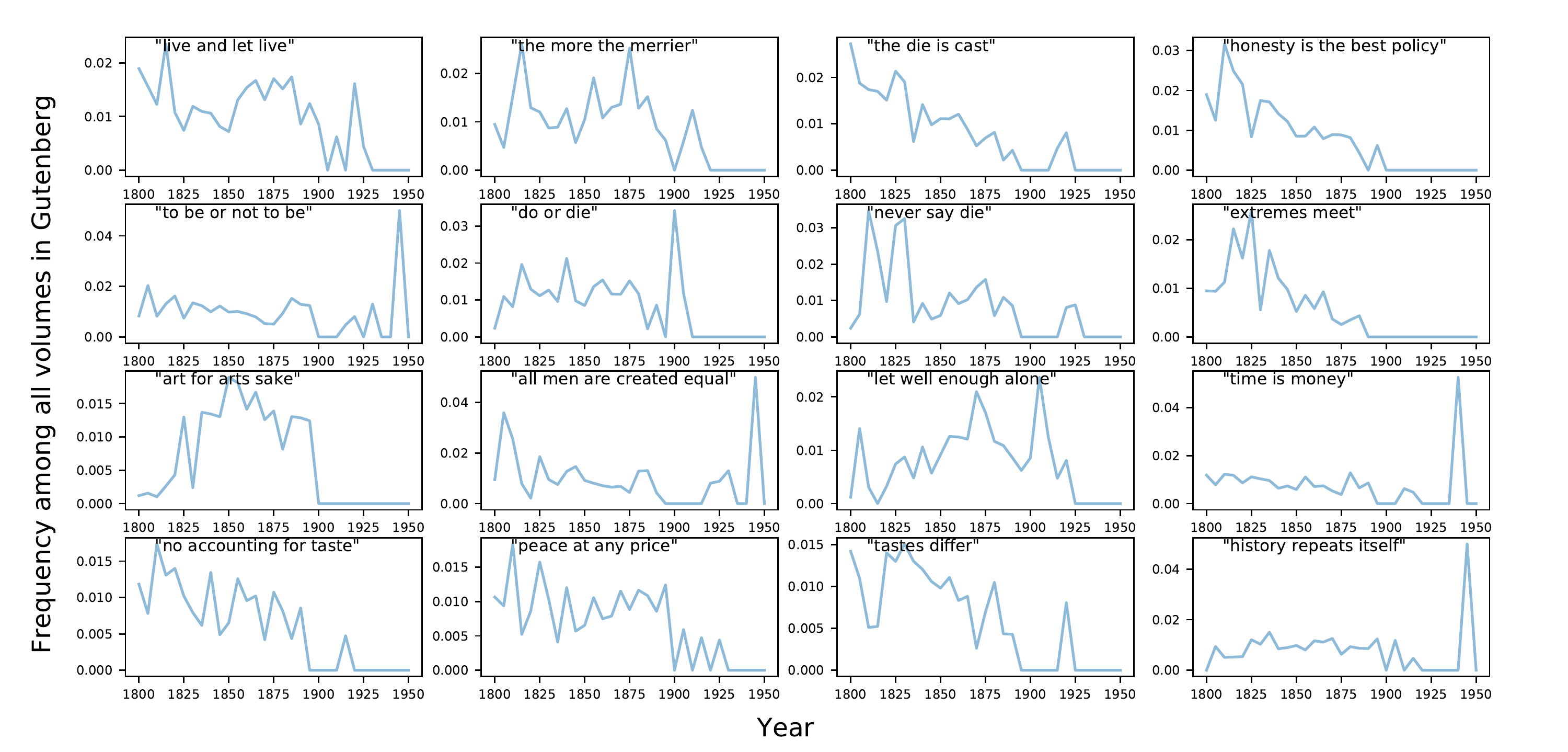}
\caption{\textbf{Time Series plots for the 17--32 most popular proverbs in Gutenberg (ranked by overall count) by 20 year bins.}}

\centering
\includegraphics[width=\textwidth]{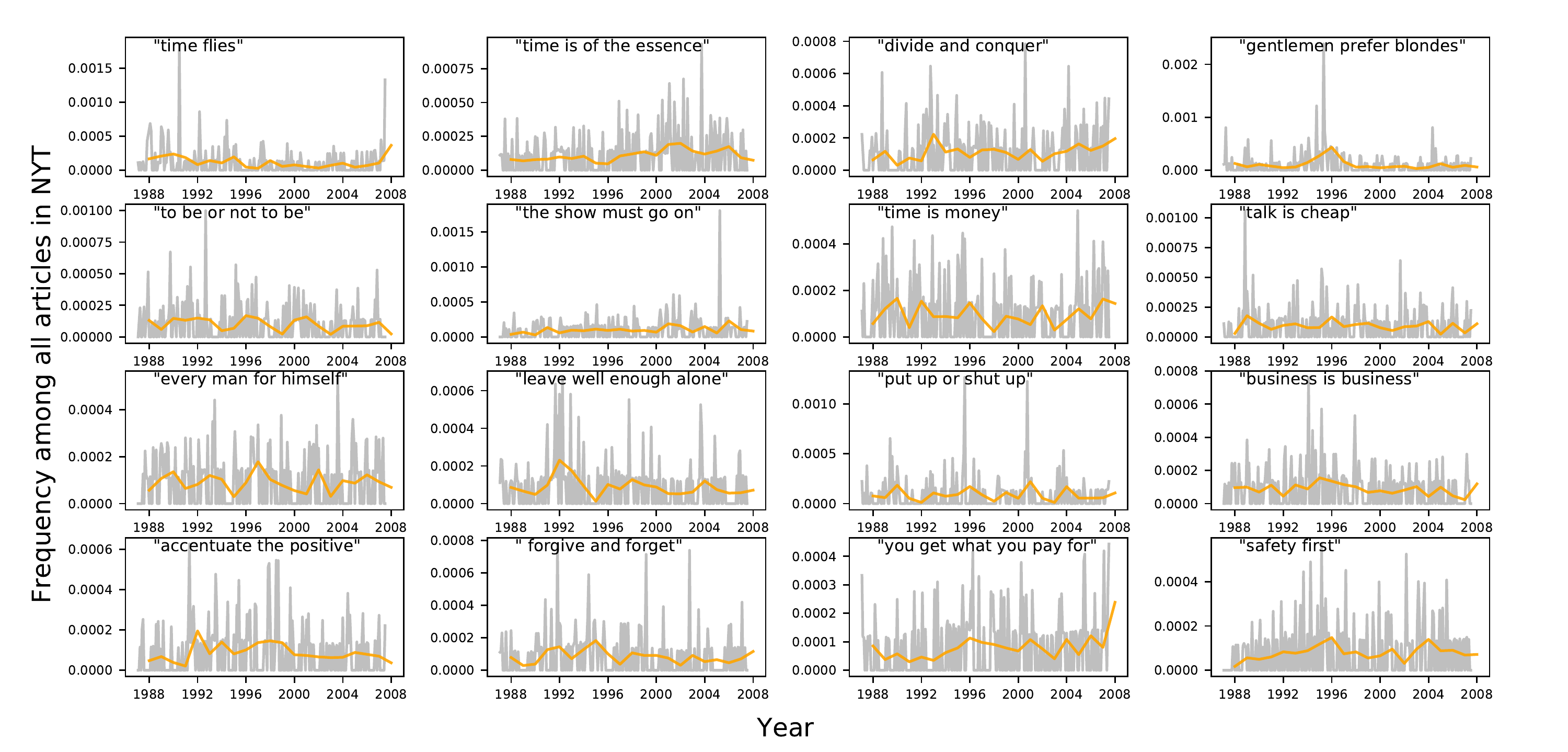}
\caption{\textbf{Time Series plots for the 17--32 most popular proverbs in \textit{The New York Times} (ranked by overall count).} The gray represent the data binned by month, and the orange represent the data binned by year.}
\end{figure*}

\begin{figure*}[h!]
\centering
\includegraphics[width=\textwidth]{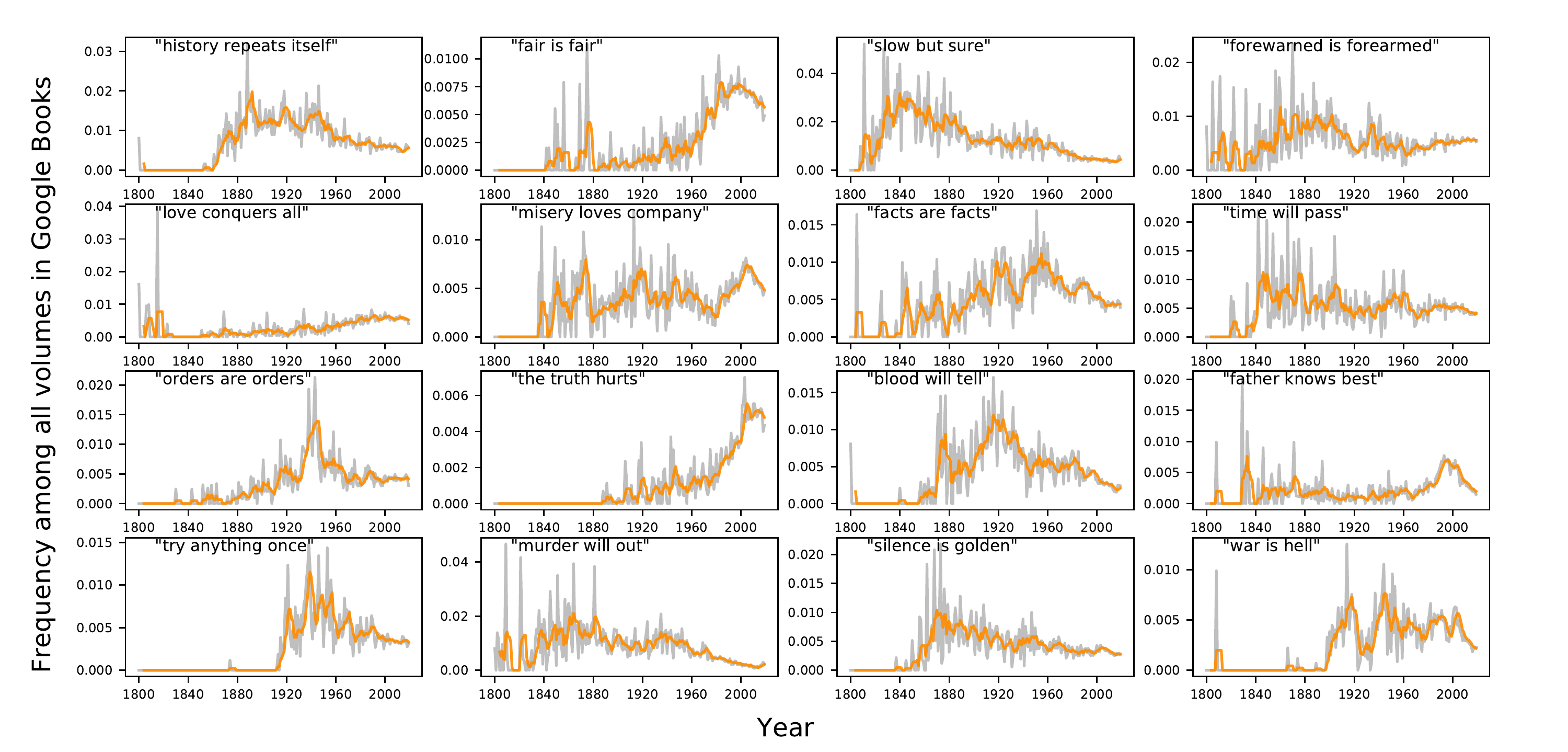}
\caption{\textbf{Time Series plots for the 17--32 most popular 3-gram proverbs in the Google Books N-gram Corpus (ranked by overall count).} The gray represents the yearly frequency, while the orange represents the 5 year rolling average.}

\centering
\includegraphics[width=\textwidth]{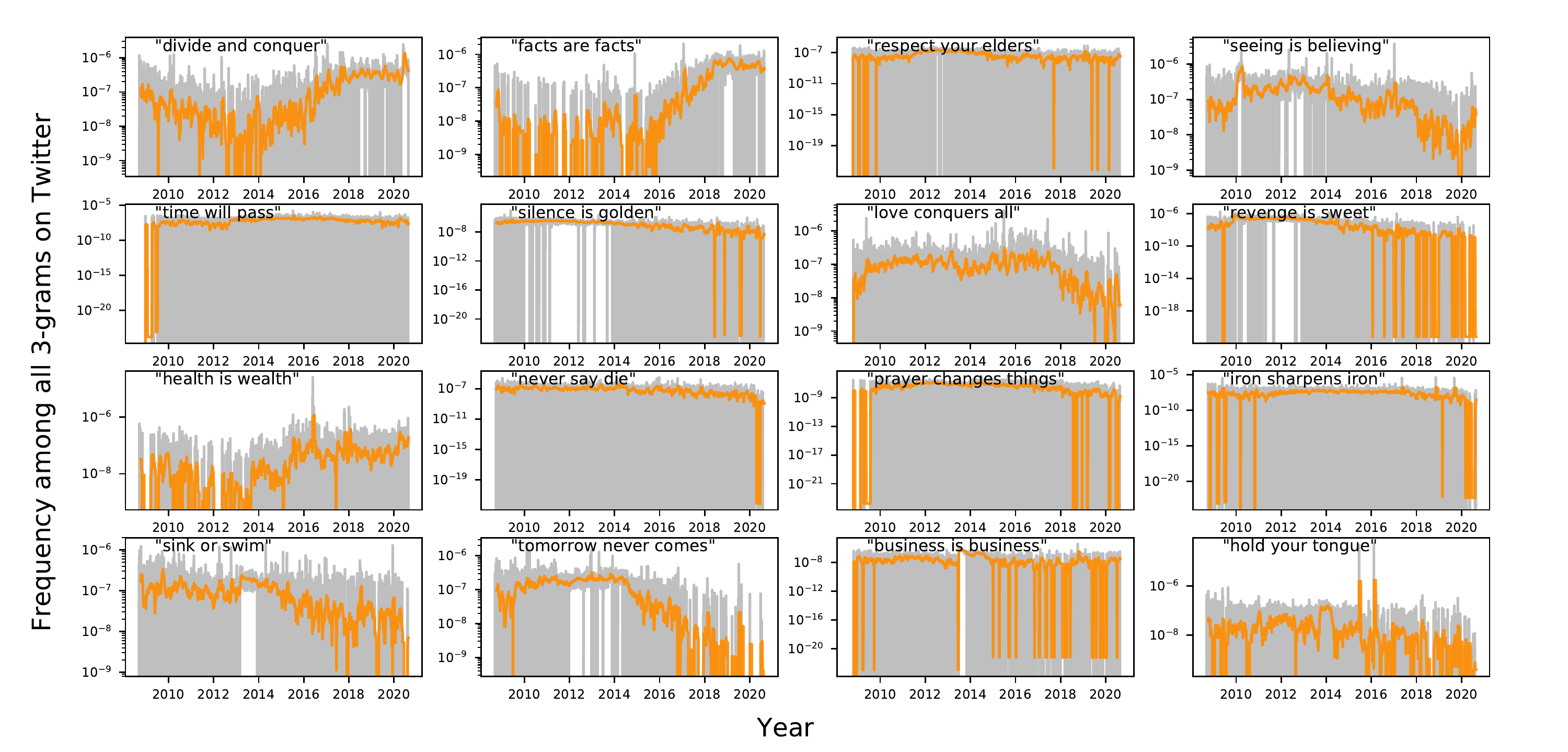}
\caption{\textbf{Time Series plots for the 17--32 most popular proverbs on Twitter (ranked by overall count).} The gray represents the daily frequency, while the orange represents the 30 day rolling average.}
\end{figure*}

\end{document}